\documentclass[a4paper,fleqn]{cas-dc}

\usepackage[numbers]{natbib}
\usepackage{adjustbox}
\usepackage{multirow}
\usepackage{multicol}
\usepackage[table]{xcolor}%
\usepackage{amssymb}
\usepackage{amsmath}
\usepackage{hyperref}
\usepackage{caption}
\usepackage{booktabs}
\usepackage{stfloats}
\usepackage{float}
\def\tsc#1{\csdef{#1}{\textsc{\lowercase{#1}}\xspace}}
\tsc{WGM}
\tsc{QE}
\tsc{EP}
\tsc{PMS}
\tsc{BEC}
\tsc{DE}

\begin{document}
\let\WriteBookmarks\relax
\def\floatpagepagefraction{1}
\def\textpagefraction{.001}

\shorttitle{Z. Zhao et~al.}

\shortauthors{Z. Zhao et~al.}

\title [mode = title]{VGS-ATD: Robust Distributed Learning for Multi-Label Medical Image Classification Under Heterogeneous and Imbalanced Conditions}                      


%
\author[1,2]{Zehui Zhao}[orcid=0000-0002-9578-5706]
\ead{zehui.zhao@hdr.qut.edu.au}


\author[2,3]{Laith Alzubaidi}
\cormark[1]
\ead{l.alzubaidi@qut.edu.au}
\author[3]{Haider Alwzwazy}
\ead{haider alwzwazy@hdr.qut.edu.au}
\author[1,2]{Jinglan Zhang}
\ead{jinglan.zhang@qut.edu.au}
\author[4]{Clinton Fookes}
\ead{c.fookes@qut.edu.au}
\author[3]{Yuantong Gu}
\ead{yuantong.gu@qut.edu.au}

\affiliation[1]{
    addressline={School of Computer Science, Queensland University of Technology}, 
    city={Brisbane},
    postcode={4000}, 
    state={QLD},
    country={Australia}}

    \affiliation[2]{
    addressline={Centre for Data Science, Queensland University of Technology}, 
    city={Brisbane},
    postcode={4000}, 
     state={QLD},
    country={Australia}}

    \affiliation[3]{
    addressline={School of Mechanical, Medical, and Process Engineering, Queensland University of Technology}, 
    city={Brisbane},
    postcode={4000}, 
     state={QLD},
    country={Australia}}

  \affiliation[4]{
    addressline={Signal Processing, Artificial Intelligence and Vision Technologies (SAIVT), Queensland University of Technology}, 
    city={Brisbane},
    postcode={4000}, 
     state={QLD},
    country={Australia}}



\cortext[cor1]{Corresponding author}
\begin{abstract}
In recent years, advanced deep learning architectures have shown strong performance in medical imaging tasks. However, the traditional centralized learning paradigm poses serious privacy risks as all data is collected and trained on a single server. To mitigate this challenge, decentralized approaches such as federated learning and swarm learning have emerged, allowing model training on local nodes while sharing only model weights. While these methods enhance privacy, they struggle with heterogeneous and imbalanced data and suffer from inefficiencies due to frequent communication and the aggregation of weights. More critically, the dynamic and complex nature of clinical environments demands scalable AI systems capable of continuously learning from diverse modalities and multilabels. Yet, both centralized and decentralized models are prone to catastrophic forgetting during system expansion, often requiring full model retraining to incorporate new data. To address these limitations, we propose VGS-ATD (Vision Transformer-based Generalized and Scalable AI-To-Data), a novel distributed learning framework. It leverages fully-connected peer-to-peer (P2P) communication to ensure privacy without relying on a central server. It conducts a one-time weight aggregation between nodes to maintain domain-specific knowledge, thereby reducing the aggregation and communication burden and preventing catastrophic forgetting. To validate VGS-ATD, we evaluate it on highly heterogeneous and imbalanced medical datasets encompassing a wide range of imaging modalities and classification tasks. In experiments spanning 30 datasets and 80 independent labels across distributed nodes, VGS-ATD achieved an overall accuracy of 92.7\%, outperforming centralized learning (84.9\%) and swarm learning (72.9\%), while federated learning failed under these conditions due to high requirements on computational resources. VGS-ATD also demonstrated strong scalability, with only 1\% drop in accuracy on existing nodes after expansion, compared to a 20\% drop in centralized learning, highlighting its resilience to catastrophic forgetting. Additionally, it reduced computational costs by up to 50\% relative to both centralized and swarm learning, confirming its superior efficiency and scalability.
\end{abstract}




\maketitle
\section{Introduction}

Nowadays, the development of AI technology has facilitated the emergence of numerous advanced architectures, further promoting their applications in real-world environments \cite{beddiar2023deep}. In the healthcare domain, AI-supported medical diagnosis has long been a topic of interest and has demonstrated promising performance \cite{beddiar2023deep,liu2025adaptive}. However, the statistical heterogeneity and privacy concerns remain crucial challenges in medical imaging learning. The differences in disease symptoms, regional variations, and imaging techniques may lead to heterogeneity issues, which significantly hinder the generalizability and performance of the medical model. Normally, using traditional Centralized Learning (CL) methods to collect and train with a large dataset can improve the model's generalizability and mitigate the negative influence of data heterogeneity. However, such learning methods have a severe risk of data leakage during model transferring and storage, which has prevented their utilization in many privacy-sensitive areas, especially in the medical field \cite{beltran2023decentralized}. 

To reduce the privacy risk and enable secure sharing between clinics, recent research has introduced two novel decentralized learning methods: Federated Learning (FL) \citep{mcmahan2017communication} and Swarm Learning (SL) \citep{warnat2021swarm, saldanha2022swarm}, providing a viable option for training models without gathering data centrally. These methods train models locally on client devices and aggregate only the models' weights to share learned knowledge directly without transferring raw data. Specifically, FL uses a central server to communicate with client nodes and conduct weight aggregation to achieve a more powerful and generalized global model \cite{banabilah2022federated}. In contrast, the SL method employs a sequential P2P model update process, passing trained model weights from a previous client to a new client and conducting model aggregation with the node's local model \cite{shammar2024swarm}. However, both FL and SL methods have several limitations, including the inability to handle data heterogeneity, high communication costs during model aggregation, catastrophic forgetting after updating the global model, and high computational resource requirements for training on edge devices \cite{rauniyar2023federated}. Their solutions aggregate a global model during the node local model training process, without supporting a free-in and free-out mode for clients who want to enter or exit the system. Moreover, the sequential training process of SL may also compromise the robustness of the entire system, as a single node's compromise or malicious behavior can affect the integrity of the entire system.

Recently, some novel frameworks have been designed to address the aforementioned challenges, providing holistic solutions that simultaneously tackle multiple limitations within a unified architecture. For example, Chakshu’s Orbital Learning framework \citep{chakshu2024orbital} integrates Split Learning \citep{vepakomma2018split} and ensemble learning to protect data privacy and enhance model performance. It demonstrates improved accuracy over traditional FL methods and supports datasets with varying feature spaces. However, its approach to data heterogeneity is limited, as it merely employs a similarity-based mechanism to avoid aggregating models from nodes with highly divergent features. Similarly, Thapa et al. proposed SplitFed \citep{thapa2022splitfed}, a hybrid framework that combines the advantages of split learning and FL. This approach mitigates the sequential dependency inherent in traditional split learning and has shown promising performance across several benchmark medical datasets. Nonetheless, their work does not account for complex real-world scenarios such as multi-label classification.

\begin{figure*}[!ht]
\centerline{\includegraphics[width=0.95\textwidth]{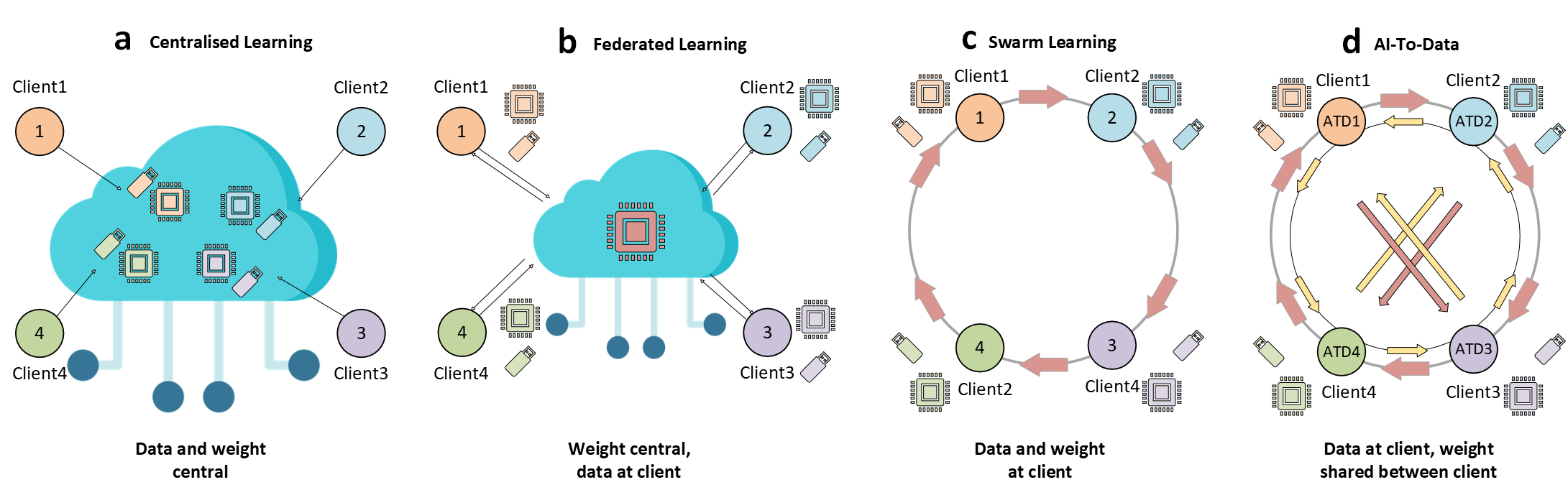}}
\caption{a, Sample of cloud-based centralized learning paradigm, each client's data and learned weights are stored in the central server. b, Sample of federated learning paradigm, as only learned weights are shared in the central server. Raw data and computational resources are kept by clients. c, Sample of swarm learning process, with both data and weight kept locally with clients. d, Sample of ATD learning process, the client can train a local model and share learned weights with each other, and keep the raw data locally. Unlike others, ATD operates efficiently at both intra-node and inter-node levels: within each node, data is split into low-resource batches (trained on single or multi-GPU setups) and trained into a local model. These local models are then shared in a fully connected manner, enabling collaboration without central coordination. ATD supports continuous, incremental learning as new data or tasks emerge, while preserving privacy and minimizing costly retraining.} 
\label{structure_comparison}
\end{figure*}

In this study, we propose a novel learning framework: Vision Transformer-based Generalized and Scalable AI-To-Data (VGS-ATD), which utilizes the AI-To-Data (ATD) \citep{alzubaidi2025atd} learning method to address several challenges, including data privacy, data heterogeneity and imbalance, and computational efficiency, as well as system adaptability and scalability. The ATD is a highly distributed P2P decentralized learning method, which allows nodes to exchange local model weights and develop a customized global model to handle different scenarios. As shown in Fig.~\ref{structure_comparison}, the primary difference between CL, FL, SL, and ATD lies in the system structure used, as ATD enables participating nodes to directly connect with other nodes, allowing for more flexible weight exchange and dynamic collaboration between clients. Benefiting from ATD's highly distributed structure, the proposed VGS-ATD framework conducts a one-time aggregation within each client by exchanging only the model's body (extractor) weight to build a global extractor first, followed by building a global classification head (classifier) using extracted CLS token embeddings through machine learning classifier training and ensemble learning. 

In VGS-ATD, we employ Vision Transformer (ViT) \citep{dosovitskiy2020image} as the foundational model, replacing commonly used Convolutional Neural Networks (CNNs). ViT is a novel deep learning architecture that leverages self-attention mechanisms to capture long-range dependencies and global contextual information. Compared to CNNs, ViT exhibits better performance in complex and high-variance medical imaging tasks, but requires a larger dataset for training to ensure satisfactory performance. Moreover, our method introduces a fully decentralized and distributed learning framework that allows clients to join dynamically without incurring additional overhead. We applied a one-time weight aggregation after local model training finishes, which enables them to learn personalized local feature information to against data heterogeneity. The follow-up feature selection, classifier training, and ensemble learning process leverages multiple robust classifiers, enabling reliable predictions and reducing the influence of feature bias and class imbalance.

To validate the effectiveness of our approach, we evaluate it under a simulated, large-scale medical environment featuring highly imbalanced and heterogeneous datasets. These datasets are derived from diverse medical imaging modalities and span up to 80 disease classes, reflecting real-world variability. We also benchmark our framework against centralized learning and decentralized learning paradigms (federated learning and swarm learning) mentioned above to better demonstrate the benefits of our design.

The main contribution of this study is listed as follows:
\begin{enumerate}
    \item We propose VGS-ATD, a novel privacy-preserved decentralized learning framework designed to build generalizable, adaptable, scalable, and computationally efficient medical diagnostic systems within a unified architecture.
    \item We introduce a comprehensive design of three VGS-ATD configurations, enabling flexible scalability and improved computational efficiency during node expansion.
    \item We conduct a three-phase empirical evaluation of VGS-ATD against three widely used learning methods on heterogeneous, imbalanced, and multi-modality medical datasets. VGS-ATD achieved up 92.7\% accuracy in the final 80-class classification task, with up to 20\% higher accuracy and 50\% lower computational cost when compared to centralized and decentralized baseline methods. The comparison between node expansion also shows its superior capability against catastrophic forgetting.
\end{enumerate}

\section{Backgrounds}
Dealing with complex and changeable medical imaging-related tasks remains a challenge to DL models. Multiple issues, including data heterogeneity, data scarcity, privacy concerns, and class imbalance, have hindered the model's ability to learn powerful feature representations from the target data. In this section, we introduce several centralized and decentralized methods that appear in the proposed method or as baselines, along with the one-time aggregation technique we apply for instant model aggregation. 

\subsection{Centralized Learning Method}
Centralized learning is one of the earliest and most widely adopted paradigms in machine learning, where all data is collected and processed on a central server. Since the success of large-scale CNN architectures such as AlexNet \citep{krizhevsky2017imagenet} and Xception \citep{chollet2017xception}, the CL method has become widely used for training deep learning models. By combining all data and computational resources, the CL method can maximize training effectiveness and data usability, ensuring satisfactory performance, especially with data-hungry architectures. In the medical imaging field, CL methods have shown promising results in maintaining high-level model performance and robustness. However, in a real-world scenario, the availability of data in a single hospital is limited due to the high expense of collecting and annotating it. Uploading medical datasets from hospitals worldwide would require numerous permissions. Moreover, the various requirements from different governments for storing and utilizing trained models make it impossible to create such an environment for conducting CL \citep{bao2022federated}. Another major challenge is the constraints of computational resources in training with a large amount of medical samples, as large-scale DL architectures can be resource-intensive and time-consuming \citep{bellavista2021decentralised}. Current DL models lack the capability to handle dispersed and heterogeneous medical data, and combining different medical datasets may also lead to differences in feature distribution, causing bias \citep{yue2020deep}. So far, a widely applied solution is to finetune part of the model when facing new disease data, which wastes computational resources and is time-consuming. Additionally, most clinical data are collected and stored independently within hospital servers, and uploading or exchanging raw data poses a high risk of data leakage. 

\subsection{Federated Learning Method}
Federated Learning \citep{mcmahan2017communication,chen2025prototype,sultan2025federated} has become a widely adopted paradigm in medical image analysis due to its strong privacy-preserving properties. FL typically falls into three main categories: 1. Horizontal federated learning (HFL): HFL is defined for client datasets that share a homogeneous feature space, but are distinct in the sample ID space. 2. Vertical federated learning (VFL): VFL deals with scenarios where two datasets share the same sample identity space, but differ concerning their feature spaces. 3. Federated transfer learning (FTL): FTL is employed when two datasets vary within both sample ID space and feature space, using transfer learning to connect different clients' datasets \citep{guendouzi2023systematic}. HFL has gained early popularity for its simplicity and ability to train a global model by aggregating local updates, without sharing raw data. All three approaches offer privacy protection by exchanging only model parameters instead of sensitive data. However, a major limitation of FL lies in its difficulty in handling heterogeneous data distributions, which is a common scenario in real-world applications. This heterogeneity can arise from differences in statistical distributions across participants \citep{gao2022survey}, or from domain shift, where local datasets vary significantly in modality, structure, or class imbalance \citep{stacke2020measuring}. Such discrepancies often lead to degraded model performance and hinder generalization across clients. 

To address the data heterogeneity problem, several modified FL structures have been proposed, including Federated Representation Learning (FRL) \citep{jing2023exploring} and FedBABU \citep{oh2021fedbabu}. The primary difference between these methods and general FL methods lies in their decomposition of the entire network into two components: the body (extractor), which is related to universality, and the head (classifier), which is related to personalization. The weight aggregation would only happen at the extractor level, and each client node trains a personalized classifier after receiving the aggregated global extractor from the central server. In this way, FRL and FedBABU achieved success in allowing the decentralized FL paradigm to learn from clients with heterogeneous feature spaces and distinct sample ID spaces. Nevertheless, such methods require additional training resources on the client side and do not allow any new client node to instantly participate or quit the system, as it necessitates the cooperation of all remaining clients to produce a new aggregated extractor.  

\subsection{Swarm Learning Method }
Swarm learning \citep{warnat2021swarm, saldanha2022swarm} emerges as a decentralized paradigm that promotes collaborative learning through P2P interaction, utilizing edge computing and blockchain technology. In contrast to many existing federated learning models, SL's P2P structure surpasses them, eliminating the need for a central server. Moreover, the blockchain technology offers strong guarantees regarding data privacy and secure communication among the nodes \citep{fragkou2024joint}. Warnat's work \citep{warnat2021swarm}, which utilizes SL in medical imaging tasks, achieved comparable performance in chest X-ray and blood cell classification tasks compared to centralized learning techniques. So far, SL has demonstrated its potential in addressing the limitations of centralized data storage and model aggregation. While SL provides a robust defense against adversarial attacks, it comes at a high computational expense. Another main limitation of SL is that its sequential P2P structure is inherently time-consuming, which does not support parallel learning like FL. Even worse, the whole system may fail in a chain reaction if a single node becomes compromised or acts maliciously.  

\subsection{AI To Data Method }

AI-To-Data \citep{alzubaidi2025atd} is a newly proposed decentralized learning method developed to address the common challenges mentioned in FL and SL. As AI systems are increasingly involved in medical applications, the need for a more powerful, generalized, scalable, stable, and privacy-preserved next-generation medical AI systems becomes urgent. As Table~\ref{comparison_CL_FL_SL} shows, existing methods like CL, FL, and SL have limitations in handling large-scale multi-label medical classification tasks. On the other hand, the ATD method has demonstrated promising potential in addressing these challenges due to its flexible and modular design. The client nodes within the ATD system function as both local and global nodes, and can exchange trained weights directly without going through a central server. ATD applied a P2P blockchain technique to ensure security during node interaction and communication. Benefit from P2P communication, each node can aggregate weights locally and build a customized global model to follow its personalized needs. Moreover, ATD supports a flexible training option with both parallel and sequential approaches, as the weight aggregation in ATD occurs after the local node model training is complete. Under this structure, each client node can train a local model without interrupting other nodes and also utilize the weights shared from other nodes to perform transfer learning, thereby achieving higher performance and boosting training efficiency. For models trained on different feature spaces and sample identity spaces, ATD enables them to aggregate only the backbone extractors and train a global classifier for prediction. Compared to the FL method, ATD eliminated the need for a central server, reducing the cost of node-central server communication and the risk of privacy leakage during communication. In contrast to SL, a more flexible P2P communication between nodes facilitates model training in a more resilient and flexible manner, ensuring the system's stability and avoiding the chain failure reaction in SL. So far, the effectiveness of ATD has only been proven in surveillance videos \citep{jebur2025scalable}, using CNNs as the base models; more domains and architectures need to be tested to further boost its development. 

\begin{table}[!ht]
\centering
\caption{An abstract comparison of Centralized learning (CL), Federated learning (FL), Swarm learning (SL), and AI to Data (ATD). }
\begin{adjustbox}{width=0.48\textwidth}
\begin{tabular}{c c c c c}
\hline
 & { CL}& { FL} & { SL} & { ATD}\\
\hline
{ Model aggregation} &{ No }&{ Yes }&{ Yes}&{ Yes}\\
{ Data privacy protection} &{ No }&{ Yes }&{ Yes}&{ Yes}\\
{ Communication protection} &{ No }&{ No }&{ Yes}&{ Yes}\\
{ Distributed computing} &{ No }&{ Yes }&{ Yes}&{ Yes}\\
{ Access to raw data} &{ Yes }&{ No }&{ No}&{ No}\\
{ P2P communication} &{ No }&{ No }&{ Yes}&{ Yes}\\
{ Client-side training} &{ No }&{ Parallel }&{ Sequential}&{ Parall \& Sequen }\\
{ Communication rounds} &{ No }&{ Multiple }&{ Multiple}&{ One Time }\\
\hline
\end{tabular}
\end{adjustbox}
\label{comparison_CL_FL_SL}
\end{table} 

\subsection{One-time Weight Aggregation}

One-time weight aggregation is an emerging and promising research direction in decentralized learning that significantly reduces communication costs. In recent FL studies, this approach attracts much attention and is often referred to as one-shot federated learning. For example, Liu et al. proposed FedLPA \citep{liu2024fedlpa}, where clients train models locally and compute the local covariance of their datasets using a layer-wise Laplace approximation. The server then aggregates these contributions to estimate the global expectation and update the global model parameters in a single aggregation step. Similarly, Zeng et al. introduced IntactOFL \citep{zeng2024one}, which integrates well-trained local models into an ensemble model and uses it as a teacher to guide a student global model. The ensemble, implemented as a mixture-of-experts network, is strengthened with self-supervised generated data and frozen to preserve local knowledge and prevent catastrophic forgetting. This ensemble then guides the training of a high-performance global gating network. Experimental results from both works show that one-time aggregation not only reduces communication overhead but can also outperform methods relying on multiple aggregation rounds, by better capturing global parameters while preserving local knowledge. In this study, we adopt a similar idea by training local models independently and then aggregating them once through P2P node communication, where all parameters are used to maintain local knowledge.

\section{Methodology}

\begin{figure*}[!ht]
\centerline{\includegraphics[width=0.95\textwidth]{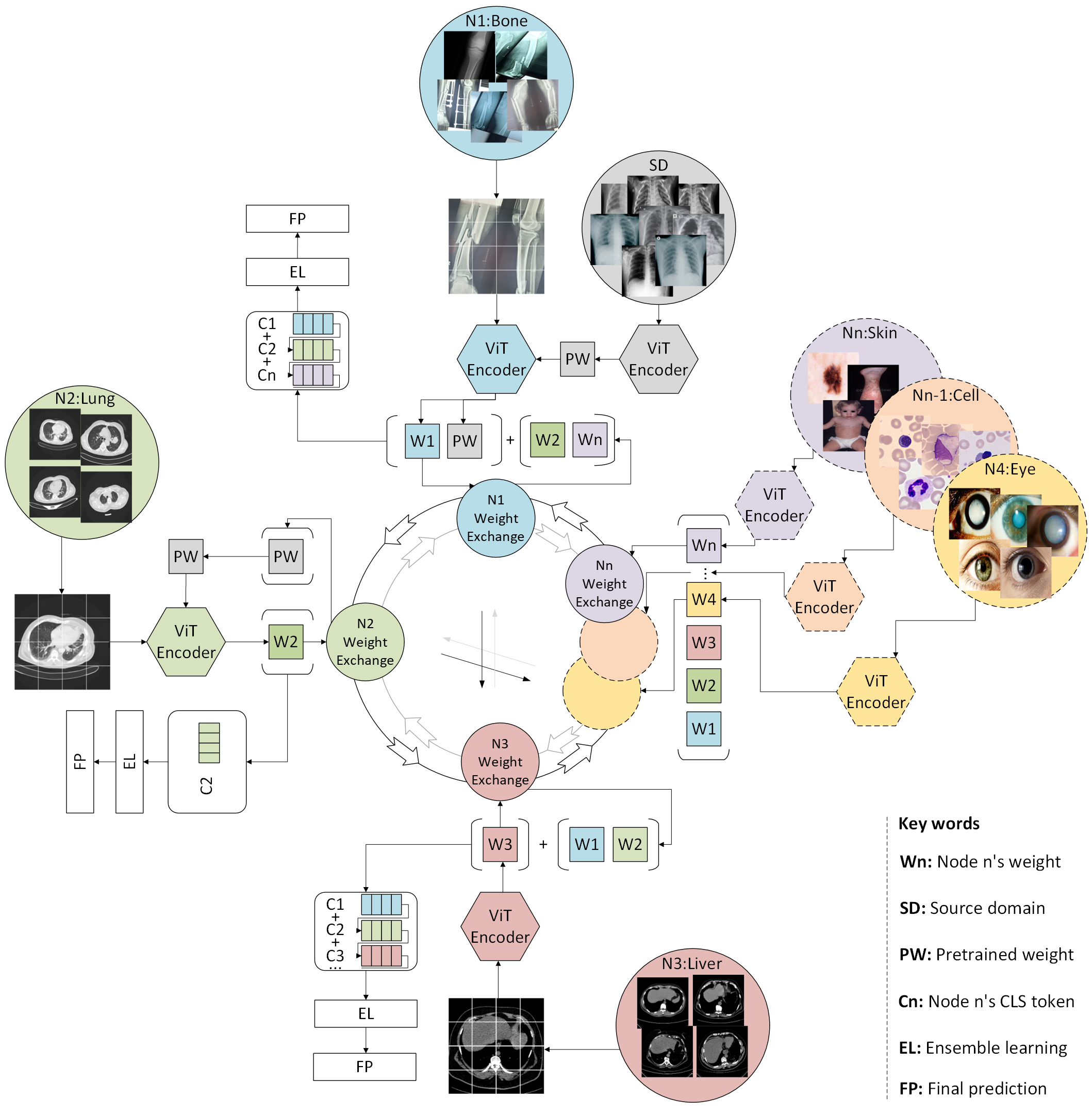}}
\caption{A sample of the VGS-ATD workflow. Each client node can train a local model using the ViT architecture and receive the backbone extractor weights from other nodes to aggregate a global extractor, utilizing ensemble learning to build a customized classifier. The pretrained weights can also be transferred through weight exchange, benefiting the local model's training efficiency and improving performance.} 
\label{Overall_workflow}
\end{figure*}

The proposed VGS-ATD method has four distinctive steps necessary for its components to function properly. As shown in Fig.~\ref{Overall_workflow}, the first step involves training a local model with the client. This step utilizes the client's local data and computational resources to maintain a personalized local model and protect the privacy of the raw data. The second step begins by excluding the classification head from each client-side model and exchanging the weight of the backbone extractor through blockchain-protected node-to-node communication. In this step, each client node sends its local extractor weight and receives weights from other verified nodes through P2P communication. The third step allows each client to build a customized global extractor using one-time weight aggregation. The client has the flexibility to select weights that contain target disease information or utilize all the received weights to build a more generalized extractor tailored to a wider group of diseases. The final step would focus on creating a new classification head using the extracted features. Benefiting from the third step, the client node now has a more generalized and powerful extractor, which it can use to extract features from local data. The extracted features are first filtered using a feature selection method, Principal Component Analysis (PCA), and then fed to train multiple downstream machine learning classifiers, including Multilayer Perceptron (MLP), Random Forest (RF), and Support Vector Machine (SVM). The predicted probability maps of each classifier are ensembled using a soft voting classifier to provide more accurate and robust final predictions. Moreover, the classifier building is not constrained within a single node, as each client node can stack together by exchanging and concatenating the CLS token extracted from the local ViT encoder block to build a global classifier that can predict a massive number of classes at once without sharing their raw data. 

In the following subsections, we present three configurations of the VGS-ATD, each tailored to address different practical medical scenarios. 

\begin{figure*}[!ht]
\centerline{\includegraphics[width=0.95\textwidth]{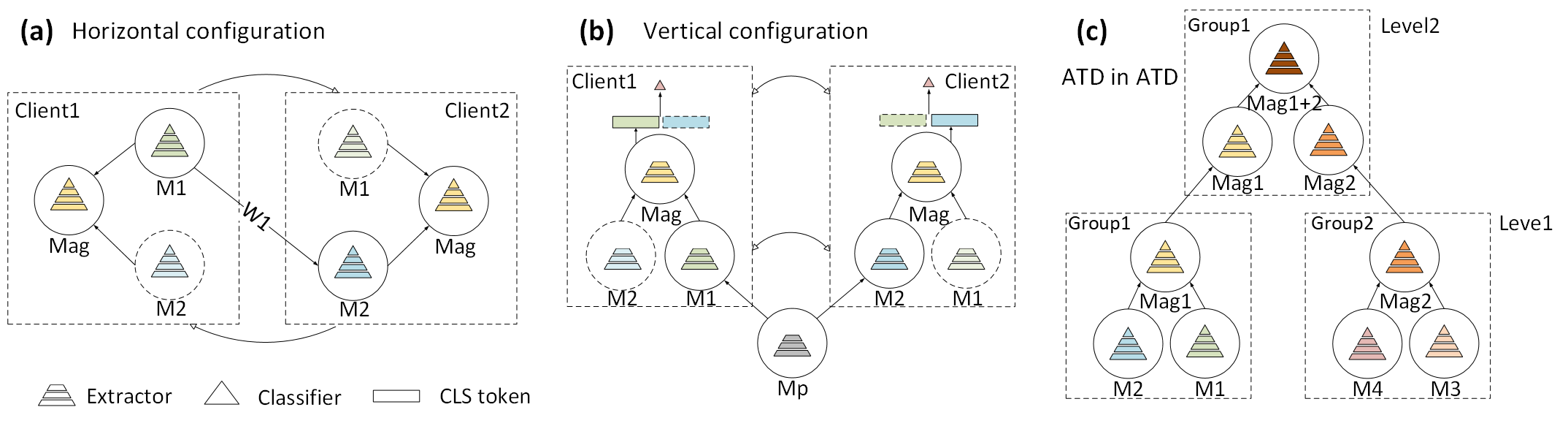}}
\caption{(a), A sample of the VGS-ATD's horizontal training configuration when two nodes share the same feature space. Mn represents the corresponding model n, and Wn represents the weights transferred from model n. The final aggregated model is called Mag. (b), A sample of the VGS-ATD's vertical training configuration when two nodes do not have overlap in feature space. Mp represents a shared pretrained model, and only the extractor part is transferred and aggregated in this setting. (c), A sample of the VGS-ATD's ATD in ATD configuration. Model aggregation can be performed in multiple level.} 
\label{Configuration}
\end{figure*}

\subsection{Configuration of Horizontal VGS-ATD}
This is the basic configuration of the VGS-ATD, which aims to expand the training dataset across client nodes without sharing raw data by using a decentralized paradigm. The workflow can be referenced in Fig.~\ref{Configuration} (a). The goal of this configuration is to utilize a broader range of distributed data while minimizing the administrative burden associated with data access permissions. It is best suited for scenarios in which client nodes share a common feature space or have overlapping feature distributions. Specifically, we conduct a node-to-node knowledge transfer learning between client nodes to enable a better local model performance and avoid domain mismatch. Specifically, an initial client node $n_1$ trains a local ViT encoder $f^1_\theta$ on its private dataset $D^{(1)} = {(x_1, y_1)}$ by minimizing the standard cross-entropy loss:

\begin{equation}
L^{(1)} = \mathbb{E}_{(x,y)\sim D^{(1)}}[l(f^1_\theta(x),y)]
\end{equation}

The remaining client nodes $n_i \in {2, \dots, N}$ initialize their local ViT encoders by loading the received backbone extractor weights $W^{(1)}$, then attach new classification heads compatible with the label space of their own private datasets $D^{(i)} = {(x_k, y_k)}$. Each client subsequently fine-tunes its ViT model $f^i_\theta$ by freezing a subset of the backbone layers and updating the remaining trainable parameters using the same cross-entropy loss as in Equation (1). No raw data or labels are shared between nodes during this process. The first trained model thus acts as a pre-trained model, enabling most of the client nodes to benefit from transfer learning.

After all client nodes have locally fine-tuned their models, they enter a second communication phase to share their updated model weights. Each node’s specific model weights $W^{(i)}$ are then aggregated by averaging to form a global model, following a strategy similar to federated learning:

\begin{equation}
    W_{fused} = \frac{1}{N}\sum^N_{i=1}W^i
\end{equation}

At this point, each client can adopt the averaged backbone weights $W_{\text{fused}}$ to construct a unified global encoder $f_{\text{fused}}$. Only structurally consistent components, such as patch embedding layers and encoder blocks, are included in the aggregation. Classification heads are excluded unless two nodes share the same label space and feature structure.

In scenarios where full model aggregation is possible, each client obtains a complete global model through this one-time update and can immediately use it for local prediction. In cases where nodes only share an overlapping feature space, aggregation is limited to the backbone encoder. These clients must perform additional steps, such as extracting CLS token representations, in order to train a new classifier tailored to their specific label space.

The global encoder is then used to extract the CLS token representations for each sample: 

\begin{equation}
    z = f_{fused}(x)[:,0]\in\mathbb{R}^{768}
\end{equation}

Here, the CLS token refers to a special learnable embedding prepended to the input sequence. It is designed to aggregate global semantic information across all image patches via Multihead Self-Attention. The resulting CLS embeddings serve as compact and expressive representations of the input, which can be used for personalized downstream classifiers or concatenated across nodes for joint feature learning. 

In the final step, the CLS-token matrix is passed to three downstream machine learning classifiers: MLP, RF, and SVM, each optimized using a different objective function:

\begin{equation}
\mathcal{L}_{\text{MLP}} = - \sum_{c=1}^{C} y_c \log \hat{y}_c, 
\end{equation}
here $\hat{y}_c$ is the predicted probability for class $c$ and $y_c$ is the one-hot encoded ground truth.
\begin{equation}
\mathcal{G}(t) = 1 - \sum_{c=1}^{C} p(c|t)^2
\end{equation}
where $p(c|t)$ is the proportion of class $c$ at node $t$.
\begin{equation}
\mathcal{L}_{\text{SVM}} = \sum{i=1}^{N} \max(0, 1 - y_i f(x_i))^2
\end{equation}
where $y_i \in \{-1, +1\}$ is the ground-truth label and $f(x_i)$ is the decision function.

The probability estimates from the three classifiers are then combined using a soft voting ensemble. Let MLP's probability map $P_{m}(y = c|x)$, RF's probability map $P_{r}(y = c|x)$, and SVM's probability map $P_{s}(y = c|x)$ denote their respective predicted probabilities for class $c$ given input data $x$. The soft voting classifier computes the average probability as:
{\small
 \begin{equation}
{P_{en}(y=c|x)=\frac{1}{3}[P_{m}(y = c|x)+P_{r}(y = c|x)+P_{s}(y = c|x)]}
\end{equation}}
The final predicted class is selected by:

\begin{equation}
    \hat{y} = \arg\max_{c} P_{en}(y = c|x)
\end{equation}

\subsection{Configuration of Vertical VGS-ATD}

The second configuration targets scenarios where client nodes have non-overlapping feature spaces, such as data from different modalities or backgrounds. As illustrated in Fig.~\ref{Configuration} (b), the development of this configuration is to enable the system to generalize across a broader range of medical conditions within a unified model, thereby reducing the need to train separate models for each disease and minimizing computational redundancy. However, transferring a model from one node to another in such heterogeneous settings can cause domain mismatch, often leading to degraded performance. This limitation occurs with many pretraining methods, particularly when the source domain differs significantly from the target domain. For example, models pretrained on general RGB datasets, such as ImageNet, often underperform on grayscale medical images \cite{wen2021rethinking}. To address this, one designated client (with sufficient computational resources) trains a model on a large-scale, publicly available medical dataset (e.g., chest X-rays). The resulting backbone extractor weights $W_p$ are then distributed to all nodes as a shared foundation. Nodes may optionally train from scratch if the pretrained weights are not suitable.

Each client node $n_i\in\{1, \dots, N\}$ receives the shared backbone weights $W_p$ and uses them to initialize its local ViT model $f^i_{\theta}$, which is trained on the private dataset $D^{(i)} = {(x_k, y_k)}$ by minimizing the same cross-entropy loss as described in Equation (1). After local training, the model is decomposed into two parts: (1) the extractor consists of patch embedding layers, encoder blocks, and CLS token; (2) the classification head comprises the projection and softmax layers. To prevent label-space conflicts across nodes, only extractor weights are exchanged among the clients.

This transfer enables each node to access complementary knowledge of representation from other domains. However, unlike the horizontal configuration, the shared global extractor $f_{\text{fused}}$ cannot be used for direct prediction due to incompatible label spaces. Instead, each client must extract feature representations from their private data and build a new downstream classifier.

Specifically, each node computes the CLS token embeddings:

\begin{equation}
z^{(i)}k = f_{\text{fused}}(x_k)[:, 0] \in \mathbb{R}^{768}, \quad \forall x_k \in D^{(i)}
\end{equation}

where $z^{(i)}_k$ denotes the CLS token embedding of sample $x_k$ from node $n_i$. These embeddings are concatenated across all nodes to form a global feature map for training a shared multi-label classifier. To reduce feature redundancy and enhance classifier performance, we apply PCA to  project the 768-dimensional CLS embeddings into a 500-dimensional subspace:

\begin{equation}
\mathbf{Z}_{\text{PCA}} = \mathbf{Z} \mathbf{W}, \quad \text{where } \mathbf{Z} \in \mathbb{R}^{N \times 768},\ \mathbf{W} \in \mathbb{R}^{768 \times 500}
\end{equation}

Here, $\mathbf{Z} \in \mathbb{R}^{N \times 768}$ is the concatenated CLS token matrix from all $N$ samples, and $\mathbf{W}$ contains the top $d$ eigenvectors of the covariance matrix of $\mathbf{Z}$. The resulting $\mathbf{Z}_{\text{PCA}}$ serves as a compact representation for global classifier training.

All nodes then train the same downstream classifiers (MLP, RF, SVM) as described in the horizontal configuration, but using the reduced PCA features as input.

\subsection{Configuration of Hierarchical ATD in ATD}

The last configuration, referred to as ATD-in-ATD (see Fig.~\ref{Configuration} (c)), aims to extend the previous two configurations by introducing a hierarchical model aggregation strategy to enhance system scalability and flexibility at minimal cost. The aim of developing this setting is to enable the integration of different node groups without discarding their previously learned knowledge. The meaning of this configuration is to address the scalability challenge in decentralized learning setups, allowing for seamless model expansion while avoiding redundant training and preserving the accumulated knowledge base. This process begins by grouping client nodes and aggregating their local models into intermediate global models using either vertical or horizontal learning. These intermediate global models are then treated as new base models and are further aggregated across groups. This hierarchical fusion strategy enables the ATD system to flexibly integrate previously trained models, allowing new incoming nodes to immediately contribute to or benefit from the collective knowledge without requiring retraining from scratch. 

Specifically, given a set of global models ${f^{(i)}_{\text{fu}}}\{i=1, \dots,N\}$ trained on datasets with potentially different numbers of classes, the system performs weighted averaging of their backbone weights. This approach prevents the weights of a base model from being excessively averaged, which could otherwise introduce bias and degrade performance on specific client nodes. The weighting can be done using two strategies:
\begin{enumerate}
    \item Square-root class-based weighting:
    Each global model’s weight is proportional to the square root of the number of classes it was trained on. Let $C_i$ denote the number of classes for model $i$. The normalized weights are:
    
    \begin{equation}  
    W_{fused} = \frac{1}{N}\sum^N_{i=1}\alpha_i W^i, \alpha_i = \frac{\sqrt{C_i}}{\sum^N_{j=1}\sqrt{C_j}}
    \end{equation}
    
    \item Uniform model count-based weighting:
    Each global model $f^{(i)}_{\text{fu}}$ is assigned a weight proportional to the number of base models it was aggregated from. Let $n_i$ denote the number of base models used to create the $i$-th global model. The normalised aggregation weights are:

    \begin{equation}  
    W_{fused} = \frac{1}{N}\sum^N_{i=1}\beta_i W^i, \beta_i = \frac{n_i}{\sum^N_{j=1}n_j}
    \end{equation}
    
\end{enumerate}

After the weight aggregation, the newly formed global model replaces the lower-level global models and is used to directly extract CLS token representations from each client node. These extracted tokens are then reduced from 768-dimensional embeddings to 500-dimensional vectors, as described in Equation (10). Finally, the reduced embeddings are used to train a group of machine learning classifiers (MLP, Random Forest, and SVM), and these classifiers are combined using a soft voting classifier to produce the final prediction.

\section{Experiment}
In this section, we conduct experiments by gradually expanding the target dataset group to simulate a complex and dynamic medical environment and test the performance change of the VGS-ATD. 
\subsection{Environment Setup}
All experiments are conducted on a single desktop setup, with datasets distributed across simulated client nodes to mimic real-world decentralized settings. The implementation is performed in Python 3.9.21, utilizing the PyTorch library (version 2.7.1 with CUDA 11.8). The desktop is equipped with an Intel i9-13900K CPU, an NVIDIA RTX 4090 GPU, and 64GB of RAM.

Each client node is configured with a Vision Transformer model using the vit\_base\_patch16\_224 architecture. The model is initially pretrained on the ImageNet-21k dataset (14 million images across 21,843 classes) and subsequently fine-tuned on the ImageNet-2012 dataset (1 million images across 1,000 classes), both at a resolution of 224×224.

In our setup, each client updates its local model for 50 training epochs. The optimizer used is AdamW, with a learning rate set to 3e-3 and a weight decay of 0.05. AdamW is chosen based on findings from the original ViT paper \citep{dosovitskiy2020image}, which demonstrated better generalization and performance than SGD and vanilla Adam optimizer. The learning rate was selected based on initial empirical observations. For example, when using a relatively large learning rate (e.g., 0.1), the model showed performance improvements during the first five epochs but then experienced a decline, suggesting overfitting due to large gradient steps. Conversely, setting the learning rate too low resulted in gradual improvement, but the model failed to converge to optimal performance within the 50-epoch limit. 


For the downstream classifier hyperparameter settings, we evaluated various configurations for each downstream classifier on the test datasets and selected the settings that yielded the best performance. Specifically, the MLP classifier was trained using the Adam optimizer for 50 epochs with a learning rate of 0.001. The SVM classifier employed the squared\_hinge loss function with a convergence tolerance of 0.01. For the Random Forest classifier, we used the Gini impurity as the splitting criterion, set max\_features to log2, and specified min\_samples\_split as 3.

\subsection{Dataset and Preprocessing}
\begin{figure*}[!ht]
\centerline{\includegraphics[width=0.95\textwidth]{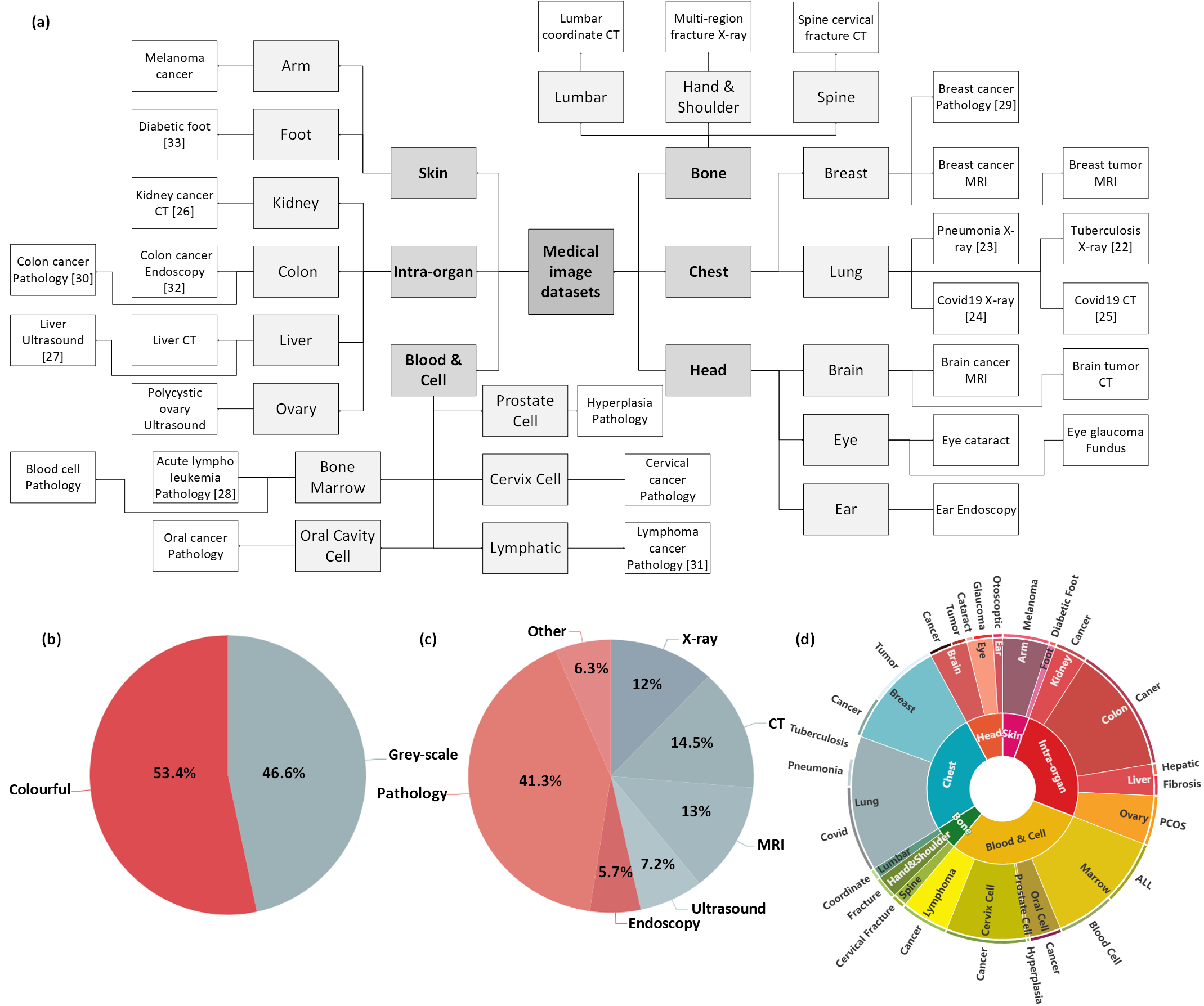}}
\caption{(a), Summary of medical datasets used in the experiment, they have been categorised based on the region of the human body and the disease type. The specific disease type and the imaging technique used to collect each dataset are also provided. (b), The ratio of grey-scale and colourful medical image samples. (c), The ratio of medical image samples collected using different imaging techniques. (d), The ratio of each dataset's samples and the body region they belong to.} 
\label{Dataset_information}
\end{figure*}

To comprehensively assess the performance of the VGS-ATD across its three configurations, we employed a diverse collection of publicly available medical imaging datasets curated from real-world clinical environments. This design aims to simulate a realistic and heterogeneous medical context. 

For the horizontal configuration test, we selected three chest X-ray datasets that focus on lung diseases and exhibit similar feature distributions: \href{https://www.kaggle.com/datasets/tawsifurrahman/tuberculosis-tb-chest-xray-dataset}{Tuberculosis Chest X-ray} Dataset \citep{rahman2020reliable}, \href{https://www.kaggle.com/datasets/paultimothymooney/chest-xray-pneumonia}{Pneumonia Chest X-ray} Dataset \citep{kermany2018identifying}, and \href{https://data.mendeley.com/datasets/8h65ywd2jr/3}{Covid Chest X-ray} Dataset \citep{shastri2022cheximagenet}. 

For the vertical configuration test, we expanded the modality coverage from chest X-rays to include four major grayscale medical imaging types: MRI, CT, X-ray, and ultrasound. This configuration incorporates 12 additional datasets spanning various anatomical regions and diagnostic tasks, including \href{https://www.kaggle.com/datasets/vuppalaadithyasairam/spine-fracture-prediction-from-xrays}{Fracture Spine} Dataset, \href{https://www.kaggle.com/datasets/bmadushanirodrigo/fracture-multi-region-x-ray-data}{Fracture Body} Dataset, \href{https://www.kaggle.com/datasets/maedemaftouni/large-covid19-ct-slice-dataset/data}{Chest Covid CT} Dataset \cite{maftouni2021robust}, \href{https://doi.org/10.34740/KAGGLE/DSV/3415848}{Kidney Cancer} Dataset \cite{islam2022vision}, \href{https://www.kaggle.com/datasets/masoudnickparvar/brain-tumor-mri-dataset}{Brain Tumor} Dataset, \href{https://www.kaggle.com/datasets/anaghachoudhari/pcos-detection-using-ultrasound-images}{PCOS Ultrasound} Dataset, \href{https://www.kaggle.com/datasets/ucimachinelearning/liver-ct-image-dataset}{Liver CT} Dataset, \href{https://www.aggle.com/datasets/murtozalikhon/brain-tumor-multimodal-image-ct-and-mri}{Brain Tumor CT} Dataset, \href{https://www.kaggle.com/datasets/vibhingupta028/liver-histopathology-fibrosis-ultrasound-images}{Liver Fibrosis Ultrasound} Dataset \citep{joo2023classification}, \href{https://www.kaggle.com/datasets/uzairkhan45/breast-cancer-patients-mris}{Breast Cancer MRI} Dataset, \href{https://www.kaggle.com/datasets/brendanartley/lumbar-coordinate-pretraining-dataset/data}{Lumbar Coordinate} Dataset, and \href{https://www.cancerimagingarchive.net/collection/breast-diagnosis/}{Breast Tumor MRI} Dataset. A total of 15 datasets are selected and stored in the corresponding simulated client nodes in this phase. This setting significantly increases the diversity and complexity of the target domain, enabling a robust evaluation of the method’s generalisability under modality shifts and data imbalance. 

Finally, for the hierarchical ATD-in-ATD configuration test, we introduce 14 color image datasets drawn from the domains of pathology, endoscopy, fundus imaging, and general clinical imaging. The aggregated global model of the color image group would then be combined with the model that we obtained from the previous horizontal configuration (which was trained with grey-scale images). The color image dataset group include the \href{https://doi.org/10.34740/KAGGLE/DSV/3415848}{Acute Lymphoblastic Leukemia} Dataset \citep{ghaderzadeh2022fast}, \href{https://doi.org/10.34740/KAGGLE/DSV/3415848}{Breast Cancer Histopathological} Dataset \citep{spanhol2015dataset}, \href{https://doi.org/10.34740/KAGGLE/DSV/3415848}{Cervical Cancer Pathology} Dataset, \href{https://doi.org/10.34740/KAGGLE/DSV/3415848}{Lung and Colon Cancer} Dataset \cite{borkowski2019lung}, \href{https://doi.org/10.34740/KAGGLE/DSV/3415848}{Lymphoma Cancer} Dataset \citep{orlov2010automatic}, \href{https://doi.org/10.34740/KAGGLE/DSV/3415848}{Oral Cancer} Dataset, \href{https://www.kaggle.com/datasets/bhaveshmittal/melanoma-cancer-dataset}{Melanoma Cancer} Dataset, \href{https://www.kaggle.com/datasets/sumithsingh/blood-cell-images-for-cancer-detection}{Blood Cell} Dataset, \href{https://www.kaggle.com/datasets/francismon/curated-colon-dataset-for-deep-learning/data}{WCE Curated Colon Disease} Dataset \citep{pogorelov2017kvasir}, \href{https://www.kaggle.com/datasets/laithjj/diabetic-foot-ulcer-dfu/data}{Diabetic Foot} Dataset \citep{alzubaidi2020dfu_qutnet}, \href{https://www.kaggle.com/datasets/nandanp6/cataract-image-dataset}{Retinal Cataract} Dataset, \href{https://www.kaggle.com/datasets/shahriar26s/benign-prostate-hyperplasiabph-detection}{Prostate Pathology} Dataset, \href{https://www.kaggle.com/datasets/ucimachinelearning/otoscopic-image-dataset}{Otoscopic} Dataset, and \href{https://pubmed.ncbi.nlm.nih.gov/21095735/}{ORIGA Retinal Glaucoma} Dataset. The number of selected datasets further increased to 29 in this phase. These datasets differ widely in image appearance, disease types, and acquisition modalities, providing an extreme-case scenario for evaluating the method’s resilience to distribution shifts, multi-label classification, and class imbalance.

Fig.~\ref{Dataset_information} presents a comprehensive overview of all datasets used in this study. Subfigure (a) provides a hierarchical summary of each dataset, organized into three levels: the general anatomical region, the specific body area where the disease occurs, and detailed dataset information including the imaging modality and reference source. Subfigure (b) illustrates the distribution of color versus grayscale images, revealing a nearly balanced composition between the two groups (approximately 50:50). Subfigure (c) highlights that the distribution of most grey-scale modalities (X-ray, CT, and MRI) is fairly balanced, with each accounting for around 13\% of the total dataset. In contrast, within the colour image datasets, 41.3\% of samples are derived from pathology images, while endoscopy and general/fundus images contribute 5.7\% and 6.3\%, respectively. Subfigure (d) complements Subfigure (a) by showing the proportional sample contribution of each dataset and its associated body region in relation to the entire dataset collection. 

The data preprocessing procedure included dataset splitting, image resizing, and target training data augmentation. To maintain consistency across datasets and facilitate reproducibility and fair evaluation, we uniformly adopted an 80:20 split ratio, while allocating 80\% of samples for training and 20\% for testing. For datasets originally provided with predefined splits by their authors, we retained the original partitioning to preserve the intended experimental conditions and enable direct comparisons with existing state-of-the-art methods built upon the same datasets.

All images were resized to 224×224 pixels to accommodate the input requirements of ViT models and to remain within the computational constraints of the available hardware. Furthermore, for datasets containing classes with fewer than 1,000 samples, data augmentation was applied to mitigate the risk of overfitting and enhance model generalizability. The augmentation techniques included random rotation ($-$30° to +30°), random horizontal flipping, zooming (scaling factor between 0.8× and 1.2×), gaussian blur ($\sigma$ = 0.1 to 2.0), and brightness adjustment (factor between 0.5 and 1.5). A custom augmentation pipeline randomly selected training samples and applied three out of these five transformations to construct the final augmented training set. 


\subsection{Evaluation Metrics}
In this study, we assessed the performance of all trained classification models using four key evaluation metrics: accuracy, precision, sensitivity, and F1-score. Accuracy measures the proportion of correctly classified samples (both positive and negative) out of the total number of samples. Precision evaluates how many of the samples predicted as positive are truly positive, and sensitivity measures the ability of the model to correctly identify actual positive cases. F1-score is the harmonic mean of precision and sensitivity, balancing the trade-off between the two. Together, these metrics provide a comprehensive evaluation of both the overall correctness of the models and their ability to correctly identify positive cases, which is especially important in medical applications.

\subsection{Baselines Setup}
To the best of our knowledge, no prior study has addressed a multilabel classification task involving such a large number of classes within a unified learning framework. Therefore, we benchmark our proposed approach against three established learning paradigms: centralized learning, federated learning, and swarm learning. This comparison aims to assess the effectiveness of each paradigm under complex and heterogeneous conditions. To ensure a fair evaluation and minimize the influence of architectural differences, all methods utilize the Vision Transformer as a common backbone.

\subsubsection{Centralized learning setup} In the centralized learning configuration, datasets from all participating nodes are aggregated into a global training pool. A unified ViT model is then trained on the combined dataset for 50 epochs using the same hyperparameter settings as the VGS-ATD. This setup serves as a performance upper bound, assuming no data leakage risk and with full data availability.

\subsubsection{Federated learning setup} The federated learning configuration adopts a state-of-the-art personalized training protocol called federated learning based on global loss decomposition (FedLD) \citep{zeng2024tackling}, which applies a novel loss decomposition to address the data heterogeneity challenge. In this setup, all base model hyperparameter setups are aligned with the VGS-ATD’s node configuration, and identical train/test splits are applied. The pretrained weights used in VGS-ATD training are also applied to initialize the base FL model. Each client performs 10 local training epochs per round, with a total of five communication rounds, ensuring that all models are trained for 50 epochs to enable a fair comparison.

\subsubsection{Swarm learning setup} The swarm learning setup follows a novel framework called non-independently and identically distributed swarm learning (Non-IID SL) \citep{gao2022new}, which is designed for decentralized heterogeneous data learning. The base model setup also mirrors those we applied for VGS-ATD and FedLD baseline, and each model is trained for 5 rounds with 10 epochs each round to ensure a fair comparison.
 
\section{Results and Analysis}
The experiments are conducted in three stages, aligned with the three configurations of the VGS-ATD.

\subsection{Phase1: Classification on 3 Nodes}
In the first phase, experiments are conducted using three datasets assigned to simulated client nodes. The results for this 3-node classification scenario are presented in Table.~\ref{Phase1_Result}. The initial three nodes each contain an independent chest X-ray dataset, with each containing a label related to normal chest and another specific chest disease label, collectively covering four unique class labels. Due to overlapping feature and label spaces in the three target nodes, this phase evaluates the performance of VGS-ATD using its horizontal configuration. To ensure a fair comparison, we apply transfer learning to all baseline methods using a shared source domain: using \href{https://www.kaggle.com/datasets/reflex7/cxr-data-set?select=Tuberculosis}{Chest X-ray} Dataset instead of ImageNet, to minimise domain mismatch risk. The selected source dataset contains approximately 72,000 chest X-ray samples spanning four lung disease categories. 

Following the initial transfer from the source dataset, VGS-ATD further performs node-to-node knowledge transfer, which is included in the horizontal configuration. However, such inter-node transfer is inherently incompatible with centralized and federated learning frameworks. In centralized learning, all datasets are aggregated and jointly trained, losing their individual identities. In the federated learning baseline, the communication structure is limited to node-to-server interactions, thereby precluding node-to-node transfer.

\begin{table}
\centering
\caption{Phase1 Performance Comparison on 3 nodes. }
\begin{adjustbox}{width=0.5\textwidth}
\begin{tabular}{c c c c c}
\hline
Method &  Acc $\uparrow$ &  Pre $\uparrow$ &  Sensi $\uparrow$ &  F1 $\uparrow$ \\
\hline
Central Learning  & 96.22\% &  96.27\% & 99.06\% & 95.69\% \\
FedLD& 83.15\%& 80.94\%& 78.32\%& 77.72\%\\
NonIID-SL & 80.31\% &  77.46\% & 70.09\% & 69.41\%\\
VGS-ATD (Ours)& \textbf{97.33\%} & 96.66\% & 97.24\% & \textbf{96.93\%}\\
\hline
\multicolumn{5}{l}{$\uparrow$: The higher the better.}
\end{tabular}
\end{adjustbox}
\label{Phase1_Result}
\end{table}

Based on the performance presented above, VGS-ATD and the baseline central learning method achieved a satisfactory overall performance, significantly surpassing the other two baselines. Moreover, VGS-ATD achieved higher accuracy and F1 score performance, suggesting that it can provide more accurate and less biased prediction results. From the test results of this phase, VGS-ATD demonstrates a good capability in combining distributed nodes and achieving a performance comparable to centralized learning when the target dataset is not highly heterogeneous and complex. 

\begin{figure}
\centerline{\includegraphics[width=0.5\textwidth]{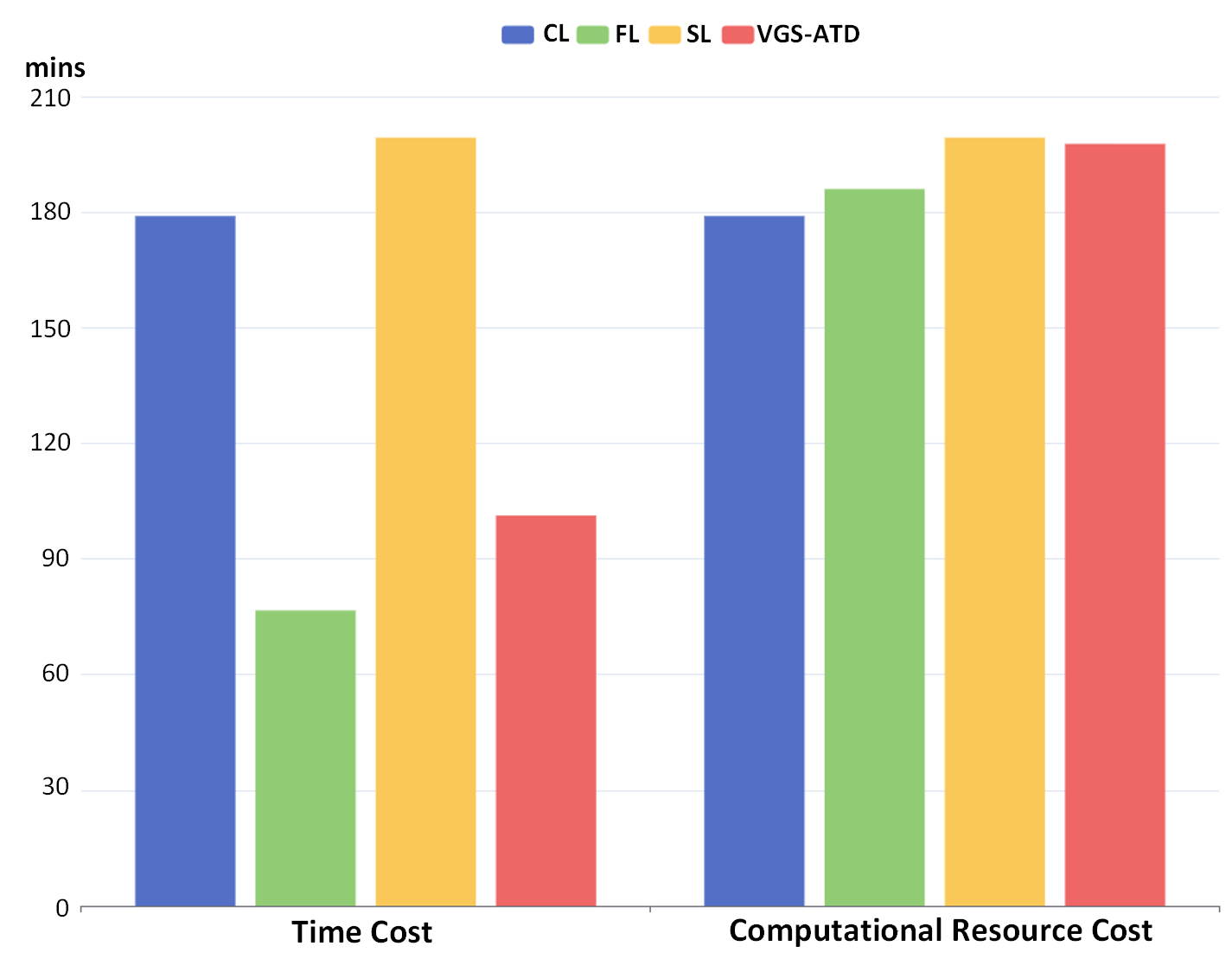}}
\caption{P1 training time and computational cost comparison.} 
\label{Phase1_Time_comparison}
\end{figure}

Additionally, Fig.~\ref{Phase1_Time_comparison} compares the training time and computational resource costs across VGS-ATD and baseline approaches. Both metrics are measured in minutes and are comparable since all methods were trained on the same hardware. Training time cost refers to the time from the start of training until the system is ready for prediction, while computational resource cost accounts for the total training time across all local models and global classifiers (as each minute spent on the same training device can be considered as a consumption of a certain amount of computational resource cost). For methods like centralized learning, which train a single model, the two metrics are effectively equivalent. In contrast, parallel approaches such as Fed-LD and VGS-ATD train local models concurrently, so their training time cost is determined by the node that takes the most time, while the overall computational cost reflects the total effort across all nodes.

As shown in the figure, the proposed and federated methods achieve significantly lower training time thanks to their parallel design. Unlike centralized and swarm learning baselines, they allow simultaneous training across nodes, which improves efficiency and scalability. For computational resource cost, centralized learning shows the best result, as it benefits from a simpler pipeline without requiring additional global classifier training. 

\subsection{Phase2: Classification on 15 Nodes}

\begin{table*}[!ht]
\centering
\caption{Phase2 performance comparison on 15-node classification task.}
\begin{adjustbox}{width=1\textwidth}
\begin{tabular}{l c c c c c c |c c c c | c c c c | c c c c}
\hline
\multirow{2}{*}{Datasets} & \multirow{2}{*}{Classes} & \multirow{2}{*}{Samples}& \multicolumn{4}{c}{Central Learning} &  \multicolumn{4}{c}{FedLD} & \multicolumn{4}{c}{NonIID-SL} & \multicolumn{4}{c}{VGS-ATD} \\
\cmidrule(lr){4-7} 
\cmidrule(lr){8-11}
\cmidrule(lr){12-15}
\cmidrule(lr){16-19}
& & & Acc &  Pre&  Sen &  F1  &  Acc  &  Pre&  Sen &  F1 & Acc  &  Pre &  Sen &  F1 &  Acc  &  Pre &  Sen &  F1\\
\hline
Chest X-Ray*&4& 21,181& 91.89& 91.89 & 91.29 & 91.59 & 75.84& 75.84 &74.62 & 75.22& 77.00& 77.00& 75.74&76.36 & \textbf{97.06}& 97.06 & 97.03& \textbf{97.04}\\

\cmidrule(lr){1-3}
\href{https://www.kaggle.com/datasets/masoudnickparvar/brain-tumor-mri-dataset}{Brain Cancer}&4&5,712 & 72.62 & 72.62 & 76.77 & 74.64 & 58.12&58.12 &68.59 &62.92 &53.17 & 53.17& 65.38& 58.65&\textbf{87.49 }& 87.49 & 87.96& \textbf{87.72} \\

\cmidrule(lr){1-3}
\href{https://www.kaggle.com/datasets/murtozalikhon/brain-tumor-multimodal-image-ct-and-mri}{Brain Tumor}&2& 3,695 &  \textbf{94.86} & 94.86 & 87.52 & 91.04 & 85.66& 85.66 &80.95 & 83.23& 86.06& 86.04& 80.51& 83.19 &92.96 &92.96 & 92.71& \textbf{92.84}\\

\cmidrule(lr){1-3}
\href{https://www.kaggle.com/datasets/uzairkhan45/breast-cancer-patients-mris}{Breast Cancer}&2& 2,080**& 69.47  & 69.47 &69.81& 69.64 & 59.62& 59.64 & 61.39& 60.49& 58.34& 57.21 & 60.10 &58.62 &  \textbf{78.37} & 78.37& 78.55& \textbf{78.46}\\

\cmidrule(lr){1-3}
\href{https://www.cancerimagingarchive.net/collection/breast-diagnosis/}{Breast Tumor}&2& 16,826&  \textbf{92.98}  &92.98  & 93.01 &  \textbf{92.99} &65.46 & 65.46 & 63.18&64.30 & 62.74&62.74 &60.49 &61.59& 92.29& 92.29& 92.21&92.25\\

\cmidrule(lr){1-3}
\href{https://www.kaggle.com/datasets/maedemaftouni/large-covid19-ct-slice-dataset/data}{Chest Covid}&3& 13,685& 91.01  & 91.01& 91.68& 91.35 & 74.28 & 74.28 & 75.83& 75.05& 72.05&72.05 &72.29& 72.17& \textbf{92.44} & 92.44&82.71 & \textbf{92.57}\\

\cmidrule(lr){1-3}
\href{https://www.kaggle.com/datasets/vuppalaadithyasairam/spine-fracture-prediction-from-xrays}{Fracture Spine}&2&3,800 & 68.25 &68.25 & 68.08 &  68.16& 74.25 &  74.25& 70.21& 72.17 & \textbf{86.26}& 86.26& 81.95& \textbf{84.04}& 80.25& 80.25& 80.05&80.15\\

\cmidrule(lr){1-3}
\href{https://www.kaggle.com/datasets/bmadushanirodrigo/fracture-multi-region-x-ray-data}{Fracture Body}&2& 6,400& 89.72 &89.72 & 86.48 & 88.07 & 64.82& 64.82 &68.33 &66.53 & 60.67&60.67 & 62.02 & 61.34 &  \textbf{98.22}&98.22 &97.07 & \textbf{97.64}\\

\cmidrule(lr){1-3}
\href{https://doi.org/10.34740/KAGGLE/DSV/3415848}{Kidney Cancer}&2& 8,000& \textbf{97.56} &97.56 & 96.78 &  \textbf{97.17} & 75.81 & 78.94 &70.89 &73.27 &78.94 &78.94 &70.21 &74.32 &96.44 & 96.44&95.54 &95.99\\

\cmidrule(lr){1-3}
\href{https://www.kaggle.com/datasets/ucimachinelearning/liver-ct-image-dataset}{Liver CT}&2&2,846 &  \textbf{90.70}& 90.70 & 93.49 & \textbf{92.07}  & 54.56& 54.56 & 69.27& 61.04& 43.16&43.16 & 63.90& 51.52& 87.89& 87.89&89.95 &88.91\\

\cmidrule(lr){1-3}
\href{https://www.kaggle.com/datasets/vibhingupta028/liver-histopathology-fibrosis-ultrasound-images}{Liver Fibrosis}&5& 5,061& 75.59&  75.59& 75.59 & 75.59 &53.85 &53.85 & 61.51&57.43& 46.94&46.94 &53.43 &49.97 & \textbf{97.73} &97.73 &97.73 & \textbf{97.73}\\

\cmidrule(lr){1-3}
\href{https://www.kaggle.com/datasets/brendanartley/lumbar-coordinate-pretraining-dataset/data}{Lumbar Coordi}&3&3,000** &  95.02& 95.02 & 88.02 & 91.39 & \textbf{95.52} &95.52 & 75.59&84.40 &89.05 &89.05 & 71.89& 79.56& 95.02&95.02 &90.52 & \textbf{92.72}\\

\cmidrule(lr){1-3}
\href{https://www.kaggle.com/datasets/anaghachoudhari/pcos-detection-using-ultrasound-images}{PCOS Ultra}&2&12,315 &97.48 &97.48  & 98.36 & 97.92& 93.06& 93.06 & 88.53&90.74 & 91.23&91.23& 88.22& 89.70& \textbf{98.25}&98.25 & 98.61& \textbf{98.43}\\

\hline
Overall:15 Sets & 35 & 112,601 &   89.96 &  85.33 & 84.10 & 83.41 &72.74 & 70.22 & 67.98 & 67.89 & 72.14 & 67.81 & 65.56 & 64.02 & \textbf{93.96}& 92.26&92.59 &\textbf{92.15}\\
\hline
\multicolumn{19}{l}{*: the marked dataset contains three sub-datasets (Chest Pneumonia, Tuberculosis, and Covid) based on Phase 1 combination. }\\
\multicolumn{19}{l}{**: The marked dataset applied data augmentation techniques to enlarge its training group. }\\
\multicolumn{19}{l}{The Samples column represents the number of training samples for each dataset. The evaluation metrics used: Acc: Accuracy, Pre: Precision, Sen: Sensitivity, F1: F1 score.}
\end{tabular}
\end{adjustbox}
\label{Phase2_Result}
\end{table*}

In the second phase, we incorporated 12 additional datasets from various grey-scale imaging modalities to simulate a scenario where new datasets are continually introduced into the system. This expansion increases the severity of data heterogeneity and class imbalance. The resulting dataset collection comprises 15 distinct subsets, including three sets inherited from Phase 1, which cover a total of 35 independent class labels. To address these new challenges, we applied the VGS-ATD using its vertical configuration. To conserve computational resources and improve training efficiency, we reused the pretrained weights derived from the Phase 1 source dataset, since the pretrained weight is based on relevant grey-scale features, providing a useful knowledge base for the newly introduced data.

Table~\ref{Phase2_Result} presents a comprehensive performance comparison, showing both the overall classification performance across all 35 classes and the individual performance of the final global model on each dataset. As expected, all learning methods experienced a performance drop compared to phase 1 due to the increased complexity of the expanded dataset. However, the performance degradation observed in the centralized learning and VGS-ATD is significantly less severe than that of federated and swarm learning settings, highlighting their superior ability to manage heterogeneous data.

Notably, VGS-ATD achieved the best accuracy performance across seven datasets, and demonstrated comparable performance to the leading methods on the remaining datasets, while centralized learning achieved a leading position in five datasets. In contrast, both federated and swarm learning baselines achieved top performance in only one dataset. These results suggest that VGS-ATD delivers more stable and balanced performance across datasets, which is highly valuable in real-world medical applications. For instance, although centralized learning reported an overall accuracy of 90\%, its performance deteriorated on small datasets such as Breast Cancer and Fracture Spine, due to its susceptibility to class imbalance.

\begin{figure}
\centerline{\includegraphics[width=0.5\textwidth]{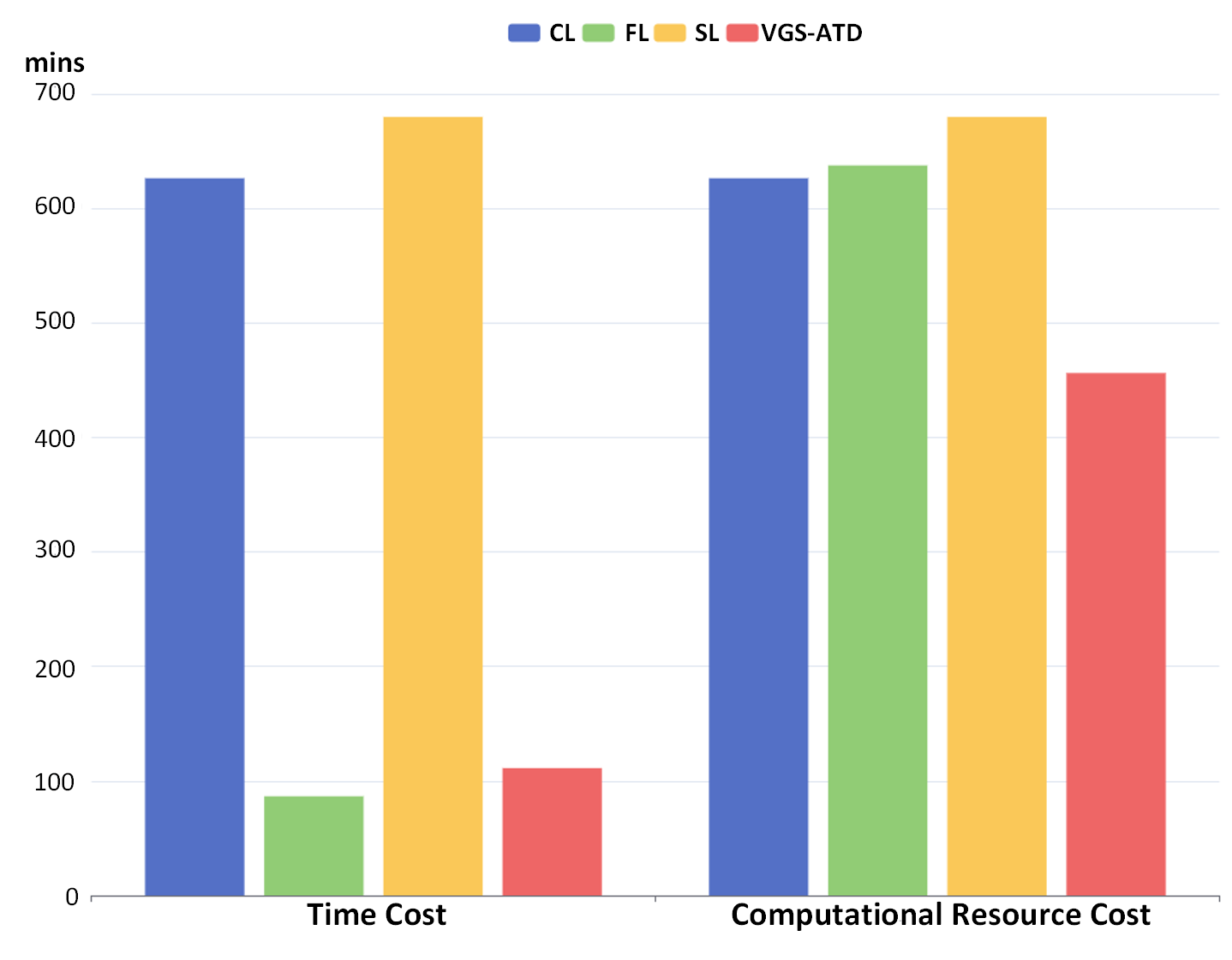}}
\caption{P2 training time and computational cost comparison.} 
\label{Phase2_Time_comparison}
\end{figure}

Another key advantage of VGS-ATD is its resilience to catastrophic forgetting. When comparing performance on the initial three Chest X-ray datasets between the two phases, all baseline methods showed a drop in accuracy. This is due to the phenomenon known as catastrophic forgetting, where central and federated learning methods require retraining the entire global model on new data, resulting in the model forgetting its previous knowledge. This issue arises because these baselines integrate all knowledge into a central model from scratch each time a new node is introduced. The reason for the performance decrease in swarm learning is different from the other two methods, mainly due to its sequential learning paradigm uses the aggregated knowledge after each round of training to initialize the next round's base node, and causes catastrophic forgetting when each round finishes. In contrast, our method preserves the previously trained weights, enabling it to retain and reuse knowledge learned in earlier phases. This ensures performance consistency across evolving data environments.

Fig.~\ref{Phase2_Time_comparison} further compares the training time and computational resource consumption across the four methods. FedLD and VGS-ATD demonstrate a clear advantage in terms of time efficiency over centralized and swarm learning. As the dataset size grows, methods using a parallel learning setting exhibit a greater advantage over those that use a sequential learning setting. 

In computational resource cost comparison, VGS-ATD demonstrates the lowest computational resource consumption. This efficiency is attributed to its scalable and distributed design, which allows it to extend its shared weight pool to incorporate new nodes without the need to retrain previously learned models. While all baselines need to be retrained with all participating nodes.

\subsection{Phase3: Classification on 29 Nodes}

\begin{table*}[!ht]
\centering
\caption{Phase3 performance comparison on 29-node classification task.}
\begin{adjustbox}{width=1\textwidth}
\begin{tabular}{l c c c c c c |c c c c | c c c c}
\hline
\multirow{2}{*}{Datasets} & \multirow{2}{*}{Classes} & \multirow{2}{*}{Samples}& \multicolumn{4}{c}{Central Learning} & \multicolumn{4}{c}{Non-IID SL} & \multicolumn{4}{c}{VGS-ATD} \\
\cmidrule(lr){4-7} 
\cmidrule(lr){8-11}
\cmidrule(lr){12-15}
& & & Acc &  Pre &  Sen &  F1  &  Acc  &  Pre &  Sen &  F1 &  Acc  &  Pre &  Sen &  F1\\
\hline

Chest X-Ray*&4& 21,181&75.98 & 75.98 &  75.56& 75.77 & 75.04&75.04  &74.09 &74.56 & \textbf{96.17}& 96.20& 96.20& \textbf{96.20}\\

\cmidrule(lr){1-3}
\href{https://www.kaggle.com/datasets/masoudnickparvar/brain-tumor-mri-dataset}{Brain Cancer}&4&5,712 & 78.41 & 78.43 & 79.44 & 78.93 & 51.41& 51.41 & 62.87 & 56.57 & \textbf{85.96}& 85.96& 86.76&\textbf{86.36} \\

\cmidrule(lr){1-3}
\href{https://www.kaggle.com/datasets/murtozalikhon/brain-tumor-multimodal-image-ct-and-mri}{Brain Tumor}&2& 3,695 & \textbf{93.91} & 93.91 & 92.66 & \textbf{93.28} &84.84 & 84.84 & 82.39& 83.60& 92.96& 92.96& 93.60& \textbf{93.28}\\

\cmidrule(lr){1-3}
\href{https://www.kaggle.com/datasets/uzairkhan45/breast-cancer-patients-mris}{Breast Cancer}&2& 2,080**& 50.48  & 50.48 &50.72& 50.60 & 54.09& 54.09 & 55.42&54.74 & \textbf{76.44}&76.44 &77.18 & \textbf{76.81}\\

\cmidrule(lr){1-3}
\href{https://www.cancerimagingarchive.net/collection/breast-diagnosis/}{Breast Tumor}&2& 16,826& \textbf{94.06}  &94.06  & 94.04 & \textbf{94.05} &60.60 & 60.60 &58.04 & 59.29& 89.62&89.63 &89.55 &89.59 \\

\cmidrule(lr){1-3}
\href{https://www.kaggle.com/datasets/maedemaftouni/large-covid19-ct-slice-dataset/data}{Chest Covid}&3& 13,685& 85.12  & 89.80& 91.92& 90.85 & 66.35& 66.35 &68.81 &67.56 & \textbf{91.56}&90.77 &91.63 &\textbf{ 91.20}\\

\cmidrule(lr){1-3}
\href{https://www.kaggle.com/datasets/vuppalaadithyasairam/spine-fracture-prediction-from-xrays}{Fracture Spine}&2&3,800 & 66.00 &65.99 & 62.56 & 68.49 & \textbf{82.75}& 82.75 & 77.70&\textbf{80.15}& 69.75& 69.75& 69.75& 69.75\\

\cmidrule(lr){1-3}
\href{https://www.kaggle.com/datasets/bmadushanirodrigo/fracture-multi-region-x-ray-data}{Fracture Body}&2& 6,400& 78.85 & 78.85& 81.93 & 80.36 &59.09& 58.10 &65.19 & 61.44& \textbf{98.61}& 96.79&96.71 &\textbf{96.75} \\

\cmidrule(lr){1-3}
\href{https://doi.org/10.34740/KAGGLE/DSV/3415848}{Kidney Cancer}&2& 8,000&\textbf{96.87} & 96.87 & 97.98 & \textbf{97.42 }&73.06 & 73.06 &67.77 & 70.32 &95.69 &95.70 & 95.03& 95.36\\

\cmidrule(lr){1-3}
\href{https://www.kaggle.com/datasets/ucimachinelearning/liver-ct-image-dataset}{Liver CT}&2&2,846 &89.64 & 89.66 & 86.46 & 88.03 & 48.94 & 48.95 &64.58 &55.69 & \textbf{90.53}& 90.53&90.53 & \textbf{90.53}\\

\cmidrule(lr){1-3}
\href{https://www.kaggle.com/datasets/vibhingupta028/liver-histopathology-fibrosis-ultrasound-images}{Liver Fibrosis}&5& 5,061& 79.25& 79.25 & 79.17& 79.21 & 47.04& 47.23& 54.69& 50.69& \textbf{97.92}&  97.92 &97.54 & \textbf{97.73}\\

\cmidrule(lr){1-3}
\href{https://www.kaggle.com/datasets/brendanartley/lumbar-coordinate-pretraining-dataset/data}{Lumbar Coordi}&3&3,000** & \textbf{97.51}& 97.52 & 93.33 & \textbf{95.38} &94.03 & 94.03 & 63.64& 75.90 &96.52 & 96.52& 90.23& 93.27\\

\cmidrule(lr){1-3}
\href{https://www.kaggle.com/datasets/anaghachoudhari/pcos-detection-using-ultrasound-images}{PCOS Ultra}&2&12,315 &91.82 &  94.39&92.78  & 93.58 & 91.64& 92.25 & 85.45& 88.72 & \textbf{97.60}&97.61 &98.24 & \textbf{97.92}\\

\cmidrule(lr){1-3}
\href{https://www.kaggle.com/datasets/mehradaria/leukemia}{ALL Cancer}& 4 & 16,000& 97.25 & 97.15  & 96.85 & 97.00 &92.06 & 92.06 &89.52 &90.77 & \textbf{98.66}&98.66 &98.38 &\textbf{98.52}\\

\cmidrule(lr){1-3}
\href{https://doi.org/10.34740/KAGGLE/DSV/3415848}{Breast Cell} &2& 8,000& 84.43  & 84.44& 90.13 & 82.80 & 82.81 & 82.81& 80.65& 81.71& \textbf{92.63}&92.63 & 93.62&\textbf{93.12}\\

\cmidrule(lr){1-3}
\href{https://www.kaggle.com/datasets/sumithsingh/blood-cell-images-for-cancer-detection}{Blood Cell}&5&16,000 &  89.88 & 89.88& 89.76& 89.82 & 90.50& 90.50 & 87.86& 89.16& \textbf{93.88}& 93.88&93.76 &\textbf{93.82}\\

\cmidrule(lr){1-3}
\href{https://doi.org/10.34740/KAGGLE/DSV/3415848}{Cervix Cancer}&5& 20,000& 88.87 & 88.88& 88.72 & 88.80 & 77.48& 77.48 & 78.69 & 78.08& \textbf{95.75}& 95.92& 96.02&\textbf{95.97}\\

\cmidrule(lr){1-3}
\href{https://doi.org/10.34740/KAGGLE/DSV/3415848}{Colon Cancer}&5&20,000 &90.55 &  92.37 &  91.36& 91.86 & 87.01&87.00 & 84.43& 85.69& \textbf{94.75}&95.01 &94.51&\textbf{94.76}\\

\cmidrule(lr){1-3}
\href{https://www.kaggle.com/datasets/francismon/curated-colon-dataset-for-deep-learning/data}{Colon Disease}&4& 12,800& 69.75 & 69.75& 70.45 & 70.10 & 71.50& 71.50  & 63.49 & 67.25 &\textbf{78.12} & 78.12&77.26 &\textbf{77.69} \\

\cmidrule(lr){1-3}
\href{https://www.kaggle.com/datasets/laithjj/diabetic-foot-ulcer-dfu/data}{Diabetic Foot}&2& 2,000**& \textbf{85.79} & 85.80  &62.50  & 72.32&49.70 & 49.70 &67.20 &57.14& 81.07& 81.07& 90.13&\textbf{85.36}\\

\cmidrule(lr){1-3}
\href{https://doi.org/10.34740/KAGGLE/DSV/3415848}{Lymph Cancer}&3& 12,000& 72.54 &	72.54  & 72.18 & 72.36 & 56.54&56.54  & 55.55&56.04 & \textbf{86.54}&86.55 & 86.43&\textbf{86.49}\\

\cmidrule(lr){1-3}
\href{https://www.kaggle.com/datasets/bhaveshmittal/melanoma-cancer-dataset}{Melanoma Cancer}&2& 11,879& 71.20 & 71.21&74.05  &  72.60& 78.25&  78.25 & 78.45& 78.35& \textbf{91.40}&91.41 & 91.17&\textbf{91.29}\\

\cmidrule(lr){1-3}
\href{https://doi.org/10.34740/KAGGLE/DSV/3415848}{Oral Cancer}&2& 8,002& 75.25 &75.24 &  74.88& 75.06 & 60.69& 60.69 &67.43 & 63.88& \textbf{80.81}& 86.55&86.43 &\textbf{86.49} \\

\cmidrule(lr){1-3}
\href{https://www.kaggle.com/datasets/ucimachinelearning/otoscopic-image-dataset}{Otoscopic}&5& 2,400&74.16 & 74.17 &74.01  &74.09  &72.50 & 72.50 &76.32 &74.36 & \textbf{96.46}& 96.47& 95.46&\textbf{95.96}\\

\cmidrule(lr){1-3}
\href{https://www.kaggle.com/datasets/shahriar26s/benign-prostate-hyperplasiabph-detection}{Prostate Cell}&2&1,000** & 81.82& 79.42 & 54.00 &  64.29 & 15.15 & 14.71 &83.33 &25.00 & \textbf{88.79}& 96.47 & 95.46&\textbf{95.96} \\

\cmidrule(lr){1-3}
\href{https://www.kaggle.com/datasets/nandanp6/cataract-image-dataset}{Retinal Cataract}&2& 2,000**& 57.69& 34.76 & 71.43 & 46.76 & 52.07& 56.28 &61.98& 58.99& \textbf{87.60}& 87.60& 87.60&\textbf{87.60}\\

\cmidrule(lr){1-3}
\href{https://pubmed.ncbi.nlm.nih.gov/21095735/}{Retinal Glaucoma}&2&4,800 & 65.94& 85.39 & 57.93 & 69.03 & 45.63 &56.18 & 63.13& 59.45& \textbf{97.19}&97.19 &97.09 &\textbf{97.14}\\
\hline
Overall: 29 Sets & 80 & 244,128&   84.92 &  82.48 & 81.27 & 78.96&73.45& 71.46& 68.39 & 67.58  & \textbf{92.64} & 90.89 & 91.35 & \textbf{90.61}\\
\hline
\multicolumn{15}{l}{*: the marked dataset contains three sub-datasets (Chest Pneumonia, Tuberculosis, and Covid) based on Phase 1 combination. }\\
\multicolumn{15}{l}{**: the marked dataset applied data augmentation techniques to enlarge its training group. }\\
\multicolumn{15}{l}{The evaluation metrics used: Acc: Accuracy, Pre: Precision, Sen: Sensitivity, F1: F1 score.}
\end{tabular}
\end{adjustbox}
\label{Phase3_Result}
\end{table*}

In the third phase, we further expanded the scope of target data by introducing a new group of color medical datasets. This color dataset group consists of 14 datasets encompassing 45 independent class labels, including pathology, fundus, endoscopy, otoscopy, and general dermatological images. The total datasets and corresponding labels in this phase reached 29 sets and 80 classes. This phase simulates a real-world scenario in which multiple independently trained dataset groups, previously using VGS-ATD, aim to be integrated to build a large platform. This scenario enables us to evaluate the performance of the ATD method in an ATD-to-ATD configuration.

In this phase, we directly applied the global model obtained from Phase 2 and performed an ensemble with a new global model trained on the incoming dataset group. Notably, compared to the earlier phases, both the sample size and the number of client nodes increased substantially, placing a heavy computational burden on our system. The centralized learning method struggled with longer data loading and computation times. In contrast, distributed learning methods, such as NonIID SL and VGS-ATD, remained unaffected, thanks to their ability to train each local model independently before aggregation. However, FedLD became impractical in this setting, as it takes significant GPU memory to run 29 independent models at the same time, and our hardware setting does not support training them simultaneously. Consequently, only centralized and swarm learning settings were included as baselines in this phase for comparative evaluation.

Another key distinction in this phase involves the choice of source domain for pretraining. Continuing to use the initial source dataset with grey-scale images would likely result in a domain mismatch when applied to the newly introduced color datasets. To address this, we utilized ImageNet pretrained weights for the new color datasets, as they provide a broader, general knowledge that includes features relevant to the human body, such as skin and eye structures. This solution enabled faster convergence and more relevant feature extraction. However, since centralized learning requires training a single unified model, it cannot accommodate two different pretrained sources. Thus, we continued to use the same source domain for the centralized learning baseline, as it performed well in Phase 2. The ImageNet weights were used exclusively with NonIID SL and VGS-ATD.

Table~\ref{Phase3_Result} reveals a consistent trend with Phase 2, showing VGS-ATD outperforming all baselines in terms of overall classification performance. As the dataset size and complexity increased, the advantages of VGS-ATD became even more pronounced. It delivered the best performance across 21 datasets and achieved comparable results to centralized learning on the remaining ones. Meanwhile, the performance of NonIID SL does not degrade after node expansion, showing the effectiveness of its design against data heterogeneity. Importantly, the initial three Chest X-ray datasets maintained stable and consistent performance with VGS-ATD across all three phases. In contrast, its performance deteriorated sharply in the other two baselines, highlighting VGS-ATD’s resilience to catastrophic forgetting.

\begin{figure}
\centerline{\includegraphics[width=0.5\textwidth]{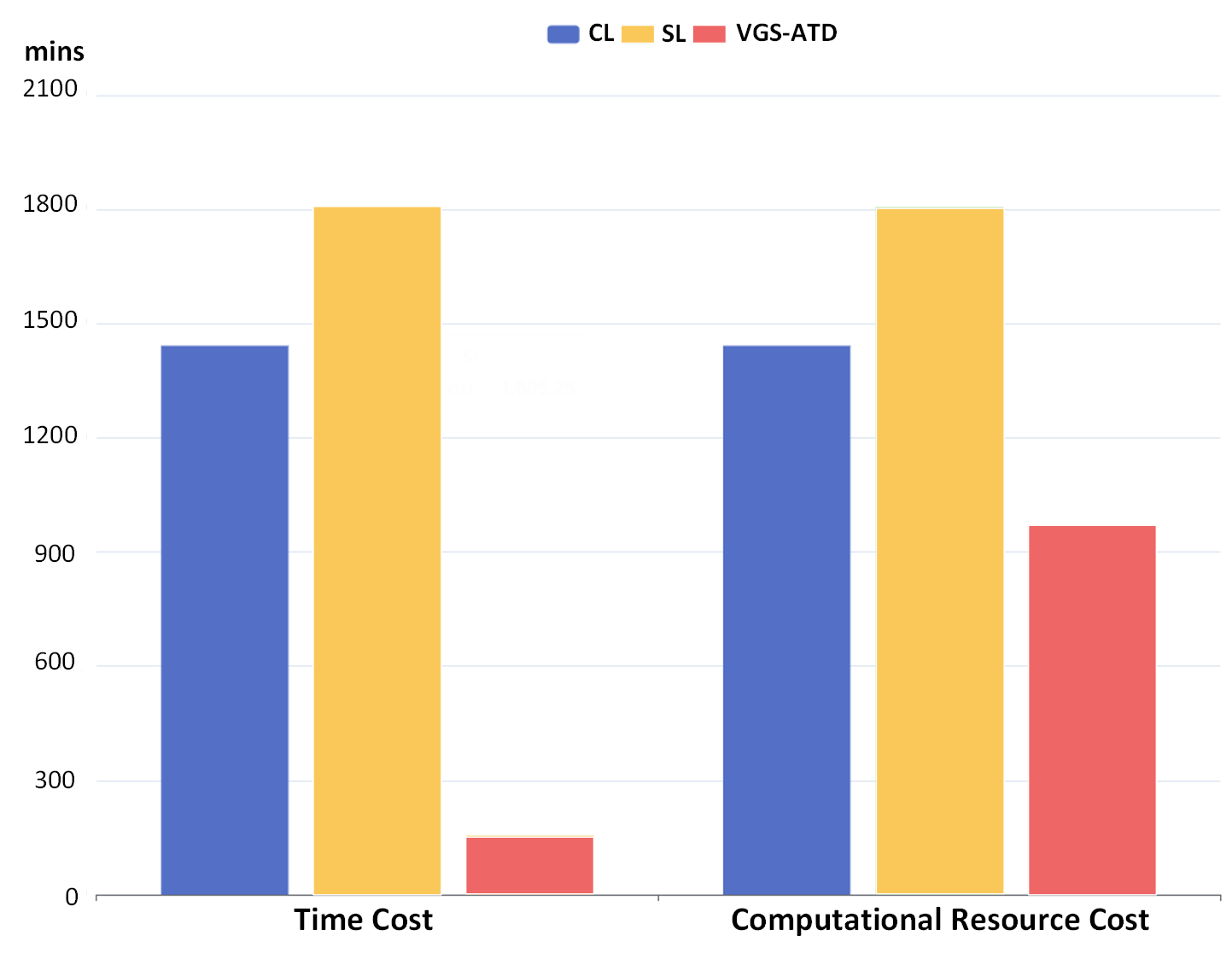}}
\caption{P3 training time and computational cost comparison.} 
\label{Phase3_Time_comparison}
\end{figure}

Fig.~\ref{Phase3_Time_comparison} illustrates the training time and computational resource consumption across methods as the dataset size increases. VGS-ATD demonstrates clear superiority in time efficiency, with the lowest overall training time among all approaches. Moreover, its computational resource cost is nearly half that of swarm learning, reinforcing its scalability and adaptability in large-scale, multi-node environments. This efficiency is attributed to its architecture, which supports incremental node integration without retraining previously trained local models.

\subsection{Robustness Analysis}
\begin{figure}
\centerline{\includegraphics[width=0.5\textwidth]{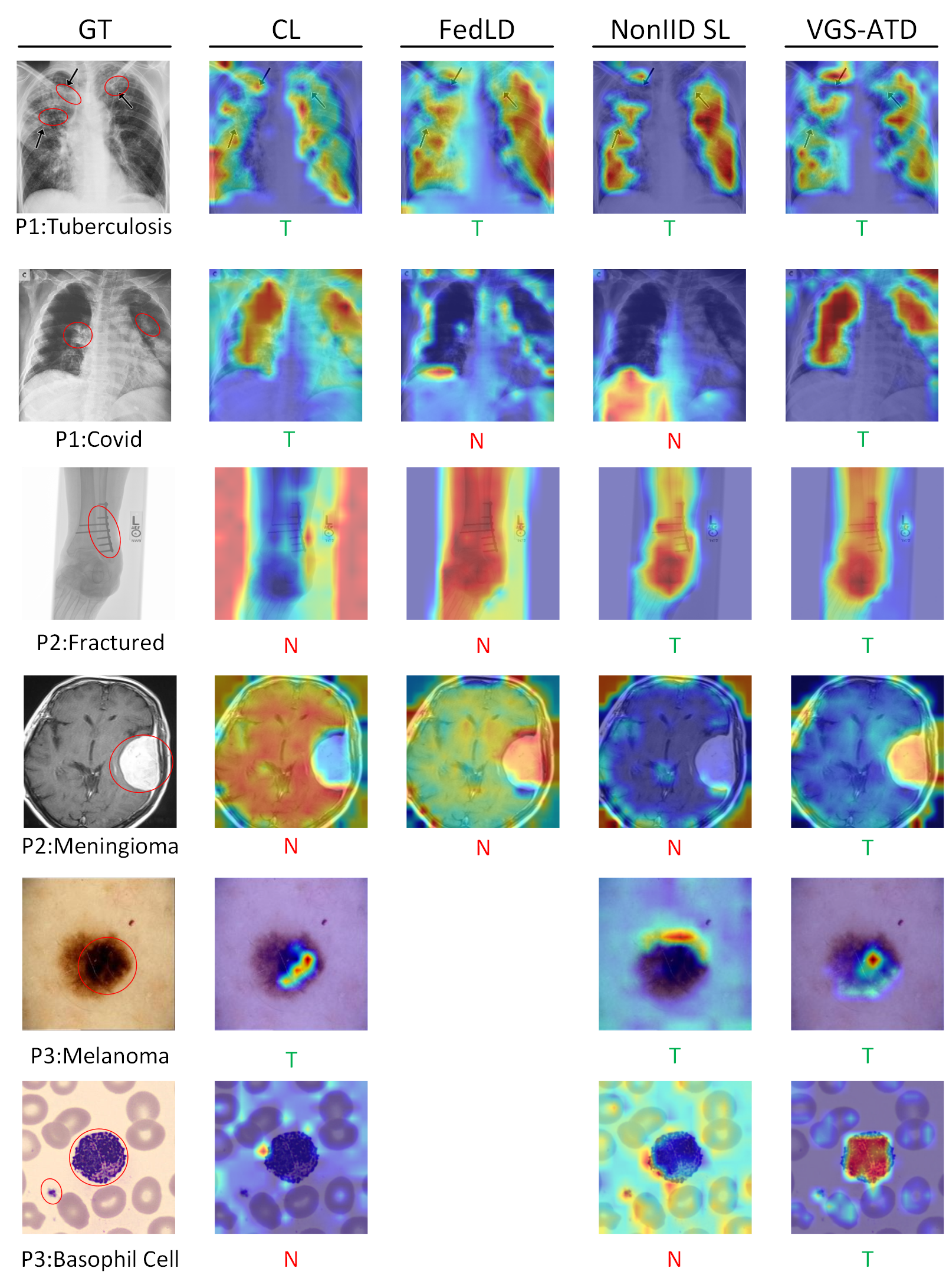}}
\caption{\textbf{Grad-CAM samples of VGS-ATD and baselines.} The title representation includes GT: Ground Truth, CL: Centralized Learning, and the other three baselines introduced above. The label representation includes P1-3: Phase 1-3, T: True for the predicted label, and N: Negative for the predicted label.} 
\label{Phase_Grad_comparison}
\end{figure}

To evaluate the robustness and reliability of VGS-ATD, we performed model attention analysis using Grad-CAM \citep{zhou2016learning}, which is a widely used Explainable AI technique. Grad-CAM integrates global average pooling to highlight the critical regions that contribute to model predictions, providing insight into decision-making. In this study, we applied Grad-CAM to the models trained by each method across all three phases of the experiment. Fig.~\ref{Phase_Grad_comparison} presents representative Grad-CAM visualization samples.

Across all phases, VGS-ATD consistently demonstrated more focused and clinically relevant attention compared to the baselines:
\begin{itemize}
    \item Phase 1: All methods correctly predicted tuberculosis by focusing on the diseased region. For the second sample (COVID-19), FedLD and NonIID SL misclassified the image, focusing on irrelevant areas of bone and soft tissue. While CL correctly localized the lung region, VGS-ATD showed a more precise focus within the lung, avoiding attention to non-pathological structures.
    \item Phase 2: At phase 2, baseline methods began to show misclassification. In the fracture sample, CL focused on background noise, and FedLD localized the bone but missed the fracture. Only NonIID SL and VGS-ATD correctly highlighted the fracture region. For the meningioma case, none of the baselines correctly localized the tumour. FedLD spread attention across the whole brain, while CL and NonIID SL failed to detect the tumor region. In contrast, VGS-ATD accurately identified the lesion.
    \item Phase 3: VGS-ATD maintained reliable performance with new color medical images. In the melanoma and basophil cell samples, it accurately localized disease-specific features. Although CL and NonIID SL produced correct melanoma predictions, NonIID SL’s attention was unfocused, casting doubt on its reliability. Moreover, both baselines failed on the basophil cell sample, misdirecting attention to background regions.
\end{itemize}

In summary, VGS-ATD not only achieved superior performance across all three phases but also delivered more interpretable and reliable attention maps. This is especially important in clinical applications, where explainability enhances trust and supports informed decision-making by healthcare professionals. 

\subsection{Ablation Study}
To assess the contribution of key components in VGS-ATD, we conducted a simple ablation study by sequentially removing four major modules: local transfer learning, node-to-node transfer learning, PCA feature selection, and classifier ensemble using soft voting. This study was conducted on a small group of nodes within the Phase 1 experiment setting to minimize computational costs. Table~\ref{Ablation study} presents the performance comparison of the global model when each component is excluded.

\begin{table}[h]
\centering
\caption{Ablation comparison on 3 nodes setting. }
\begin{adjustbox}{width=0.5\textwidth}
\begin{tabular}{c c c c c}
\hline
Settings &  Acc $\uparrow$ &  Pre $\uparrow$ &  Sensi $\uparrow$ &  F1 $\uparrow$ \\
\hline
Without Base TL  & 81.42\% &  82.44\% & 74.89\% & 76.96\% \\
Without Node-Node TL & 95.61\% & 94.72\% & 94.92\% & 94.77\%\\
Without PCA Selection& 96.89\% &  95.78\% & 95.82\% & 95.79\%\\
Without Classifier Ensemble& 94.81\% & 93.86\% & 93.35\% & 93.48\%\\
\hline
VGS-ATD (Full) & 97.33\% & 96.66\% & 97.24\% & 96.93\%\\
\hline
\multicolumn{5}{l}{$\uparrow$: The higher the better, TL represent transfer learning.}
\end{tabular}
\end{adjustbox}
\label{Ablation study}
\end{table}

The results show that removing any component results in reduced performance, confirming the effectiveness of each module. However, the degree of impact varies as follow:

\begin{itemize}
    \item Base Transfer Learning has the most significant influence. Its removal causes a significant performance drop (15\%), highlighting the importance of initializing models with a relevant source domain. It also helps mitigate data scarcity issues at the local node level, which can be another reason why removing it causes such a significant performance drop.
    \item Node-to-Node Transfer Learning leads to moderate degradation (1.5\%), as base transfer learning already provides a strong foundation in this setting. Nevertheless, node-to-node transfer can be more beneficial in cases where of domain mismatch occurs between the source and target data.
    \item PCA Feature Selection has minimal effect in this setup, likely due to the limited complexity of feature combinations in the small 3-node setting. Its benefits should become more evident in large-scale environments with diverse feature spaces.
    \item Classifier Ensemble improves robustness by aggregating predictions from multiple classifiers. Its absence reduces stability and overall accuracy, confirming the ensemble's contribution.
\end{itemize}
In summary, the ablation study underscores the importance of base transfer learning. Highlight the potential benefits from the combination of node-to-node knowledge transfer, feature refinement, and ensemble learning.

\subsection{Discussion}
Based on extensive experiments and comparisons with three mainstream learning paradigms, VGS-ATD demonstrates superior overall performance, as shown in Table.\ref{General comparison}, Its overall accuracy performance consistently outperforms state-of-the-art decentralized methods that are tailored for data heterogeneity by up to 17\%, 21\%, and 19\% accuracy across the three experiment phases. Benefited from the one-time weight aggregation, the proposed VGS-ATD also achieves up to 90\% reduction in training time and 50\% savings in computational cost compared to baselines. Additionally, by comparing the accuracy performance change in the phase 1 chest x-ray dataset, the VGS-ATD has only 1\% drop after multi-phase expansion, whereas all baselines show substantial performance degradation on previously trained nodes. The modular and scalable design of VGS-ATD allows seamless integration of new datasets and models without full retraining. In low-resource environments, its hierarchical structure enables efficient training by partitioning large datasets into manageable subsets for deployment on low-end devices. These strengths make VGS-ATD highly compatible with real-world clinical workflows and well-suited for continuous, non-disruptive learning.

\begin{table}[!ht]
\centering
\caption{Overall comparison of VGS-ATD against state-of-the-art baselines in Phase1 (3 Nodes with 4 classes), Phase2 (15 Nodes with 35 classes), and Phase3 (29 Nodes with 80 classes).}
\begin{adjustbox}{width=0.5\textwidth}
\begin{tabular}{c c c c c c}
\hline
Phase & Method &  Acc $\uparrow$ &  CXR\_Acc  $\uparrow$  &Time Cost $\downarrow$ &  Comp Cost $\downarrow$   \\
\hline
\multirow{4}{*}{Phase1} & CL & 96.22 & 96.22&  178.97 & \textbf{178.97}  \\
 & FedLD \cite{zeng2024tackling} & 83.15 & 83.15  & \textbf{76.64} & 185.98 \\
 & NonIID SL \cite{gao2022new}& 80.31 & 80.31&  199.33 & 199.33  \\
 & \textbf{VGS-ATD (ours)}& \textbf{97.33} & \textbf{97.33} &  101.26 & 197.70  \\
\hline
\multirow{4}{*}{Phase2} & CL & 89.96 & 91.89 &  766.37 & 766.37  \\
 & FedLD \cite{zeng2024tackling}& 72.74 & 75.84 &  86.64 & 637.57 \\
 & NonIID SL \cite{gao2022new}& 72.14 & 77.00 &  679.93 & 679.93  \\
 & \textbf{VGS-ATD (ours)}& \textbf{93.96} & \textbf{97.06} &  \textbf{111.26} & \textbf{456.25}   \\
\hline

\multirow{3}{*}{Phase3}& CL & 84.92 & 75.98 &  1441.96 & 1441.96  \\
 & NonIID SL \cite{gao2022new}& 73.45 & 75.04&  1805.25 & 1805.25  \\
 & \textbf{VGS-ATD (ours)}& \textbf{92.64}& \textbf{96.17}  &  \textbf{154.18} & \textbf{969.89}   \\
\hline
\multicolumn{6}{l}{$\uparrow$: The higher the better, $\downarrow$: The lower the better.}\\
\multicolumn{6}{l}{Both Acc are calculated using percentage, where CXR\_ACC represents average}\\
\multicolumn{6}{l}{accuracy on three Phase1 chest x-ray nodes.}\\
\multicolumn{6}{l}{Both computational and time costs are calculated using minutes.}
\end{tabular}
\end{adjustbox}
\label{General comparison}
\end{table}

Despite its strengths, several challenges exist. One is data scarcity at the node level. While we adopt the ViT model for its strong performance, ViT requires more training samples than CNNs. In some cases, CNNs outperform ViTs on small datasets due to better robustness. Although we applied data augmentation to help reduce overfitting, it can also increase computational cost and limit training speed. Another issue is that we tested the whole system under an ideal environment without considering real-world communication overload. Due to the resource and time limitations, we plan to conduct verification experiments in the future. The last challenge lies in the lack of inter-node contrastive learning. The above experiment suggests that VGS-ATD performs well when datasets differ in modality or feature space. However, when several nodes each contain visually similar features but represent different classes (as seen in some skin disease datasets), distinguishing between classes becomes difficult. We assume this is likely due to the one-shot aggregation strategy applied in VGS-ATD, where nodes train independently without exposure to features from other labels. Merging their weights without global fine-tuning may cause semantic confusion.

\section{Conclusion}
In this article, we propose VGS-ATD, a Vision Transformer based, generalized, and scalable distributed learning framework tailored for complex medical environments. The framework is designed to tackle key challenges in medical machine learning, including data privacy, data heterogeneity, class imbalance, computational efficiency, and scalability. We conducted comprehensive experiments to compare VGS-ATD with three widely used learning paradigms under increasingly complex clinical scenarios. Across all experiment phases, VGS-ATD demonstrated strong and consistent performance, maintaining over 92\% accuracy even when handling an 80-class classification within a unified model, and significantly outperforming both centralized and decentralized baselines. These experimental results highlight VGS-ATD’s potential as a solution for next-generation medical AI, particularly in scenarios that require robustness, adaptability, and distributed collaboration. 

In the future, we plan to evaluate the stability of the whole system by testing the nodes with real-world communication overload scenarios, and exploring a wider variety of datasets and modalities to evaluate and enhance the generalization capability of VGS-ATD in broader clinical settings. Additionally, we aim to investigate alternative backbone architectures to provide users with greater flexibility in model design.

\begin{flushleft} \large{\textbf{CRediT authorship contribution statement}} \end{flushleft}
\textbf{Zehui Zhao:~} Conceptualisation, Methodology, Validation, Writing – original draft, \textbf{Laith Alzubaidi:~} Conceptualisation, Methodology, Resources, Formal analysis, Writing – review \& editing, Validation, Supervision, Project administration, Funding acquisition, \textbf{Haider Alwzwazy:~} Formal analysis, Writing – review \& editing, \textbf{Jinglan Zhang:~}Writing – review \& editing, Formal analysis, Validation, Supervision, \textbf{Clinton Fookes:~} Writing – review \& editing, Validation, \textbf{Yuantong Gu:~} Writing – review \& editing, Validation, Funding acquisition. 
\begin{flushleft} \large{\textbf{Data statement}} \end{flushleft}
All the datasets used in this study are based on publicly available data, and the hyperlinks to these datasets are as follows: \href{https://www.kaggle.com/datasets/tawsifurrahman/tuberculosis-tb-chest-xray-dataset}{Tuberculosis Chest X-ray Dataset}, \href{https://www.kaggle.com/datasets/paultimothymooney/chest-xray-pneumonia}{Pneumonia Chest X-ray Dataset}, \href{https://data.mendeley.com/datasets/8h65ywd2jr/3}{Covid Chest X-ray Dataset}, \href{https://www.kaggle.com/datasets/vuppalaadithyasairam/spine-fracture-prediction-from-xrays}{Fracture Spine Dataset}, \href{https://www.kaggle.com/datasets/bmadushanirodrigo/fracture-multi-region-x-ray-data}{Fracture Body Dataset}, \href{https://www.kaggle.com/datasets/maedemaftouni/large-covid19-ct-slice-dataset/data}{Chest Covid CT Dataset}, \href{https://doi.org/10.34740/KAGGLE/DSV/3415848}{Kidney Cancer Dataset}, \href{https://www.kaggle.com/datasets/masoudnickparvar/brain-tumor-mri-dataset}{Brain Tumor Dataset}, \href{https://www.kaggle.com/datasets/anaghachoudhari/pcos-detection-using-ultrasound-images}{PCOS Ultrasound Dataset}, \href{https://www.kaggle.com/datasets/ucimachinelearning/liver-ct-image-dataset}{Liver CT Dataset}, \href{https://www.aggle.com/datasets/murtozalikhon/brain-tumor-multimodal-image-ct-and-mri}{Brain Tumor CT Dataset}, \href{https://www.kaggle.com/datasets/vibhingupta028/liver-histopathology-fibrosis-ultrasound-images}{Liver Fibrosis Ultrasound Dataset}, \href{https://www.kaggle.com/datasets/uzairkhan45/breast-cancer-patients-mris}{Breast Cancer MRI Dataset}, \href{https://www.kaggle.com/datasets/brendanartley/lumbar-coordinate-pretraining-dataset/data}{Lumbar Coordinate Dataset}, \href{https://www.cancerimagingarchive.net/collection/breast-diagnosis/}{Breast Tumor MRI Dataset}, \href{https://doi.org/10.34740/KAGGLE/DSV/3415848}{Acute Lymphoblastic Leukemia Dataset}, \href{https://doi.org/10.34740/KAGGLE/DSV/3415848}{Breast Cancer Histopathological Dataset}, \href{https://doi.org/10.34740/KAGGLE/DSV/3415848}{Cervical Cancer Pathology Dataset}, \href{https://doi.org/10.34740/KAGGLE/DSV/3415848}{Lung and Colon Cancer Dataset}, \href{https://doi.org/10.34740/KAGGLE/DSV/3415848}{Lymphoma Cancer Dataset}, \href{https://doi.org/10.34740/KAGGLE/DSV/3415848}{Oral Cancer Dataset}, \href{https://www.kaggle.com/datasets/bhaveshmittal/melanoma-cancer-dataset}{Melanoma Cancer} Dataset, \href{https://www.kaggle.com/datasets/sumithsingh/blood-cell-images-for-cancer-detection}{Blood Cell Dataset}, \href{https://www.kaggle.com/datasets/francismon/curated-colon-dataset-for-deep-learning/data}{WCE Curated Colon Disease Dataset}, \href{https://www.kaggle.com/datasets/laithjj/diabetic-foot-ulcer-dfu/data}{Diabetic Foot Dataset}, \href{https://www.kaggle.com/datasets/nandanp6/cataract-image-dataset}{Retinal Cataract Dataset}, \href{https://www.kaggle.com/datasets/shahriar26s/benign-prostate-hyperplasiabph-detection}{Prostate Pathology Dataset}, \href{https://www.kaggle.com/datasets/ucimachinelearning/otoscopic-image-dataset}{Otoscopic Dataset}, and \href{https://pubmed.ncbi.nlm.nih.gov/21095735/}{ORIGA Retinal Glaucoma Dataset}. 
\begin{flushleft} \large{\textbf{Consent for publication}} \end{flushleft}
All authors have given their consent for publication.
\begin{flushleft} \large{\textbf{Declaration of competing interest}} \end{flushleft}
All authors have declared that they have no conflict of interest.
\begin{flushleft} \large{\textbf{Acknowledgements}} \end{flushleft}
I would like to thank my supervisors, colleagues, and reviewers for supporting my work and providing me with valuable feedback and suggestions.
\begin{flushleft} \large{\textbf{Funding}} \end{flushleft}
Laith Alzubaidi and Yuantong Gu acknowledge the support received through the following funding schemes of the Australian Government: Australian Research Council (ARC) Industrial Transformation Training Centre (ITTC) for Joint Biomechanics under grant (IC190100020).
\bibliographystyle{elsarticle-num}
\bibliography{Reference}


\end{document}